\RequirePackage{fix-cm}
\documentclass[smallextended]{svjour3}       
\smartqed  

\usepackage{graphicx}
\usepackage[breaklinks,colorlinks,linkcolor=black,citecolor=black,urlcolor=black]{hyperref}
\usepackage{multirow}
\usepackage{subfigure}
\usepackage{rotating}
\usepackage{booktabs}
\usepackage{threeparttable}
\usepackage[numbers,sort&compress]{natbib}
\begin{document}

\title{Deep Neural Network Based Relation Extraction: \\
An Overview
}


\author{Hailin Wang
        \and
        Ke Qin \thanks{Corresponding author: Ke Qin}
        \and
        Rufai Yusuf Zakari
        \and
        Guoming Lu
        \and
        Jin Yin
}



\institute{Authors \at
              Trusted Cloud Computing and Big Data Key Laboratory of Sichuan Province, \\
              School of Computer Science and Engineering, \\
              University of Electronic Science and Technology of China, Chengdu 611731, China\\
              \email{lynn\_whl@msn.com, qinke@uestc.edu.cn, rufaig6@gmail.com, lugm@uestc.edu.cn, yinjin@wchscu.cn}           
}

\date{Received: date / Accepted: date}

\maketitle

\setcounter{tocdepth}{2}

\begin{abstract}
Knowledge is a formal way of understanding the world, providing a human-level cognition and intelligence for the next-generation artificial intelligence (AI). One of the representations of knowledge is semantic relations between entities. An effective way to automatically acquire this important knowledge, called Relation Extraction (RE), a sub-task of information extraction, plays a vital role in Natural Language Processing (NLP). Its purpose is to identify semantic relations between entities from natural language text. To date, there are several studies for RE in previous works, which have documented these techniques based on Deep Neural Networks (DNNs) become a prevailing technique in this research. Especially, the supervised and distant supervision methods based on DNNs are the most popular and reliable solutions for RE. This article 1) introduces some general concepts, and further 2) gives a comprehensive overview of DNNs in RE from two points of view: supervised RE, which attempts to improve the standard RE systems, and distant supervision RE, which adopts DNNs to design sentence encoder and de-noise method. We further 3) cover some novel methods and recent trends as well as discuss possible future research directions for this task.

\end{abstract}

\section{Introduction}
\label{Introduction}

Artificial intelligence (AI) integrating knowledge is a hot topic in current research. It provides human thinking for AI to solve complex tasks. One of the most important techniques for supporting this research is knowledge acquisition, also called relation extraction (RE). One aim of RE is to process the human language text, to find unknown relational facts from a plain text, organizing unstructured information into structured information. A well-constructed and large-scale knowledge base can be useful for many downstream applications and empower knowledge-aware models with the ability of commonsense reasoning, thereby paving the way for AI.

RE build a large-scale knowledge base by extracting relation triples from raw text. For example, there is a sentence: "\textless e1\textgreater Jobs \textless/e1\textgreater is the founder of \textless e2\textless Apple\textless/e2\textgreater." It marks the entity "Jobs" and "Apple" by a pair of XML tags. From the sentence, the RE model outputs a triple (Jobs, Apple, founded\_by), which can be used for knowledge base construction.

\begin{figure*}
\centering
\includegraphics[angle=-0,width=1.0\textwidth]{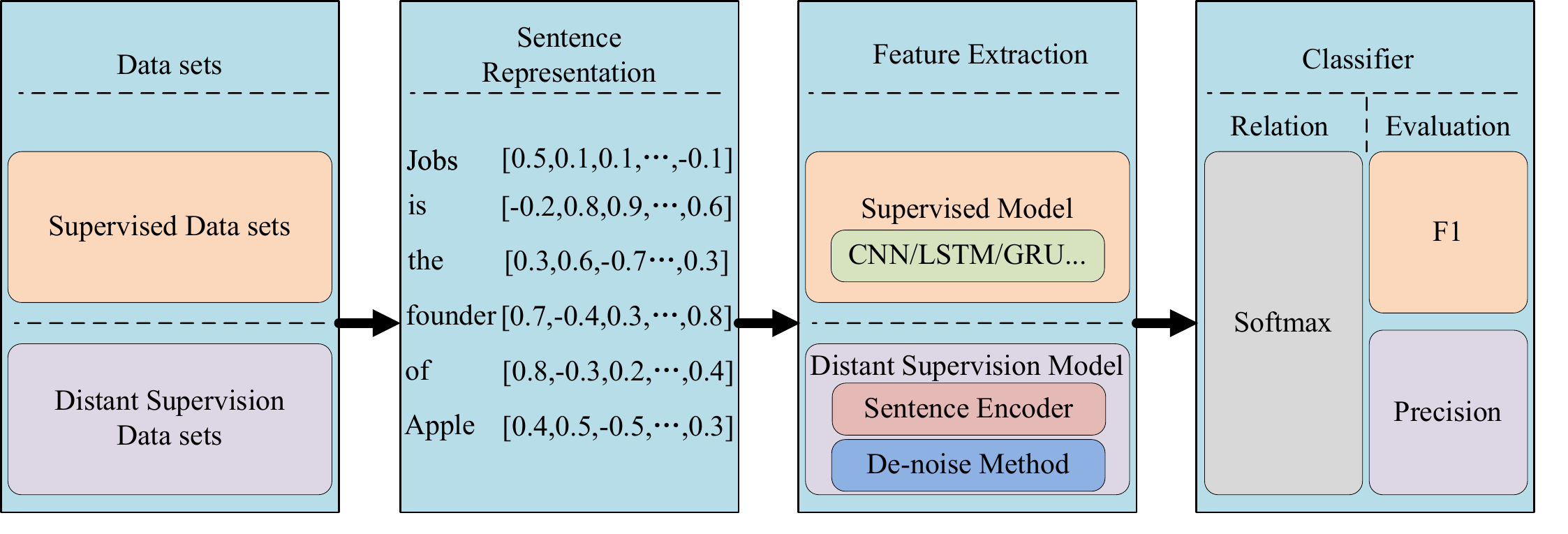}
\caption{The General Framework of DNN-based RE.}
\label{General-framework}
\end{figure*}

Recently, RE has attracted extensive attention, but few researchers to report the review of DNN-based RE \cite{kumar2017survey,pawar2017relation}. While these articles have their emphasis, lacking a comprehensive, systematic introduction to DNN-based methods.

Consequently, this paper presents an extensive survey and gives a comprehensive introduction of RE to the prevalent DNN-based methods. To begin with, this paper introduces the premise of RE frameworks, including a general framework and some basic conceptions of RE. Second, a brief introduction of traditional methods and variation of DNN-based methods will be compared in detail. Third, the paper further provides an analysis of some problems and proposes future research directions.

\section{Premise}
\label{Conceptual Framework}

\subsection{General Framework}
\label{General_Framework}

By conducting an extensive literature review, the framework of DNN-based RE methods are summarized as four components, which are shown in Figure \ref{General-framework}. The detailed description of these four components are as follows:

\textbf{Data sets}: The supervised data sets (SemEval 2010-task8 \cite{hendrickx2009semeval} and FewRel \cite{han2018fewrel,gao2019fewrel}) are often obtained by manual annotation with high accuracy and low noise, but small size. Instead, distant supervision data sets usually are acquired by a physical alignment of entities between a corpus and a knowledge base (KB), which have a bigger size and high multi-domain applicability (e.g., Riedel et al. \cite{riedel2010modeling}), but low accuracy and high noise. In a sense, this component is not a simple module in general DNN-based models. But in RE field, especially in distant supervision, the physical align phrase, constructing training triples, plays a vital role in RE.

\textbf{Sentence Representation}: In NLP field, to understand human language for computers, words are usually represented as a series of real value vectors, such as word2vec and Glove methods. Meanwhile, the position embedding is introduced to better express the positional relation between words and the entity pair. Hence, the final representation of the words is the combination of word vector and position embedding. Consequently, the final sentence representation is composed of these word representations.

\textbf{Feature extraction}: In general, these DNN-based methods are fed with the sentence representation above. With the annotated data sets, these methods output a feature extractor by training. And this extractor can extract high-level features from the above sentence representation.

\textbf{Classifier}: With the high-level features and a predefined relation inventory, the classifier outputs the relation between the entity pair in the sentence, and then evaluates the result.

\subsection{Basic Conception}

In addition to the above general frameworks, the basic concepts of DNN-based RE system commonly used in these frameworks are as follows:

\textbf{Neural Networks} have been widely used in image processing, language processing, and other fields in recent years, with very remarkable results. Researches have designed many kinds of DNNs, including Convolutional Neural Networks (CNNs) \cite{lecun1989backpropagation}, Recurrent Neural Networks (RNNs) \cite{elman1991distributed}, Recursive Neural Networks \cite{socher2012semantic}, and Graph Neural Networks (GNNs) \cite{scarselli2008graph}.
Different kinds of DNNs have different characteristics and advantages in dealing with various language tasks. For example, the CNNs, with parallel processing ability, are adept at processing local and structural information.
Instead, the RNNs, having advantages in dealing with a long text, cope with time-series information by considering the factors before and after data input.
Moreover, developed gradually in recent years, the GNNs, another kind of neural network, which processes data with a graphical structure. For example, the grammatical dependency parse tree, a general tool for RE, is suitable for the GNNs.
In addition to the above mentioned commonly used networks, there are also some other RNNs variant networks used in RE systems, such as LSTM (Long Short Term Memory network) \cite{zhang2015bidirectional,sundermeyer2012lstm,hochreiter1997long}, GRU (Gated Recurrent Unit) \cite{chung2014empirical}.

\textbf{Word Embedding}\label{word_embedding} is a method used to represent words in NLP, leveraging uniform low dimensional, continuous, real-value vectors to represent language. One of the earlier forms is one-hot, which exits some problems like data sparsity, no meaning, dimensional disasters.
To solve these problems, some scholars \cite{mikolov2013efficient} propose a new method called word2vec to overcome these disadvantages.
In this way, all the word vectors are distributed, the dimensions of the vector can be arbitrary, generally in 50 to 100 dimensions, and the value of the element can be any real value.
The greatest benefit of this approach is the semantic and contextual information of words can be captured, and the similarity of words could be calculated by simple addition and subtraction. Hence, word2vec is a common component in DNN-based RE. Aside from the word2vec method, some researchers also designed other methods \cite{turian2010word,pennington2014glove}.

\begin{table}[]
\centering
\caption{Example sentence with position indicators.}
\label{table_PI}
\begin{tabular}{cl}
\toprule
   Examples:  &\textless e1\textgreater Jobs \textless/e1\textgreater is the founder of \textless e2\textgreater Apple\textless/e2\textgreater.		              															\\
	Indictors: &\textless e1\textgreater, \textless/e1\textgreater, \textless e2\textgreater, \textless/e2\textgreater			                                                                              \\
\bottomrule
\end{tabular}
\end{table}

\begin{figure*}
\centering
\includegraphics[angle=-0,width=0.5\textwidth]{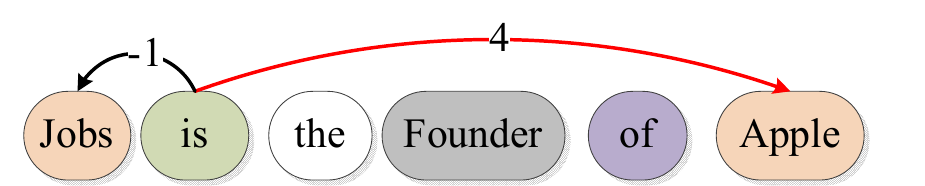}
\caption{Example of relative distance.}
\label{fig2}
\end{figure*}

\textbf{Position Embedding}\label{position_embedding} provides a uniform way for the RE model to be aware of word positions. In RE task, the CNN-based models are lack of judgment on the word location information. To address this issue, Zeng et al. \cite{zeng2014relation} propose the position feature (PF), which will be adopted in subsequent methods of using CNNs \cite{nguyen2015relation,santos2015classifying,wang2016relation}, RNNs \cite{zhang2015relation,qin2017designing,zhang2019multi}, and mixed frameworks  \cite{ren2018neural,zhang2018relation}. The PF is a combination of the relative distance between each words around the labeled entities in the sentence. For instance, given labeled entities: "Jobs" and "Apple", in the sentence "Jobs is the founder of Apple", the relative distance of the word "\emph{is}" to "\emph{Jobs}" is -1, to "\emph{Apple}" is 4. In this way, the distance of words around the entity words in the sentence can be expressed clearly. Furthermore, to make the model easy to understand the PF, the above two real values are mapped into a new vector space, namely the position embedding process. Normally the dimension of this vector is 5. One example is shown in Figure \ref{fig2}.

In addition to PE, some RNN-based methods also use position indicators (PI) to further enhance the representation of entity pair. In SemEval 2010-task8, this data set uses four position indicators to indicate the entity pair in the sentence. The example is shown in Table \ref{table_PI}.

\textbf{Shortest Dependency Path}\label{the shortest dependency path} (SDP) is a word-level de-noise method and derived from a grammatical dependency tree, which masks irrelevant words influencing the relation of entities in a sentence. (The grammatical dependency tree can be obtained by the Stanford Parser\footnote{http://nlp.stanford.edu/software/lex-parser.shtml}).
Bunescu and Mooney \cite{mooney2006subsequence} first used the SDP to design a kernel-based method to finish RE task, and then a lot of SDP-based models follow this work, such as SDP-LSTM \cite{xu2015classifying}, BRCNN \cite{cai2016bidirectional}, DesRC(BRCNN) \cite{ren2018neural}, Att-RCNN \cite{guo2019single}, and FORESTFT-DDCNN \cite{jin2020relation}.
For instance,
a sentence "\textless e1\textgreater People\textless/e1\textgreater have been moving back into \textless e2\textgreater downtown\textless/e2\textgreater." can be parsed to an SDP:
"$[People]_{e_1}$$\to$moving$\to$into$\to$ $[downtown]_{e_2}$".
This example illustrates that the SDP captures the predicate-argument sequences. In a sense, these sequences have great benefits for RE: Firstly, this method compresses the information content of sentences.
Secondly, it directly shows the dependency relations between each word. Finally, it also provides a clearer relation direction between entities.

\section{Traditional Methods}

\subsection{Categories Of Methods}
The existing RE approaches can be divided into
\textbf{
hand-built pattern methods, semi-supervised methods, supervised methods, unsupervised methods}, and
\textbf{distant supervision methods.}
In this paper, we refer to the five methods without DNNs as traditional methods.


\textbf{Hand-built pattern methods} require the cooperation between domain experts and linguists to construct a knowledge set of patterns based on words, part of speeches, or semantics.
With this linguistic knowledge and professional domain knowledge, RE can be realized by matching the preprocessed language fragment with the patterns. If they match, the statement can be said to have the relation of the corresponding pattern \cite{hearst1992automatic,berland1999finding}. Table \ref{tab1} is an example of a pattern for hyponymy.



\begin{table}[!t]
\caption{ The example of pattern "such Y as X".}
\def\arraystretch{2}
\ignorespaces
\centering
\def\arraystretch{1.15}
\begin{tabular}{ll}
\toprule
Pattern  & such Y as X\\
\midrule
Corpus   & \begin{tabular}[c]{@{}l@{}}... works by such authors as Herrick,\\ Goldsmith, and Shakespeare.\end{tabular}                                                                             \\
\midrule
Relation & \begin{tabular}[c]{@{}l@{}}Hyponym ("author", "Herrick"),\\ Hyponym ("author", "Goldsmith"),\\ Hyponym ("author", "Shakespeare")\end{tabular}
\\
\bottomrule
\label{tab1}
\end{tabular}
\end{table}

\begin{figure*}
\centering
\includegraphics[angle=-0,width=0.5\textwidth]{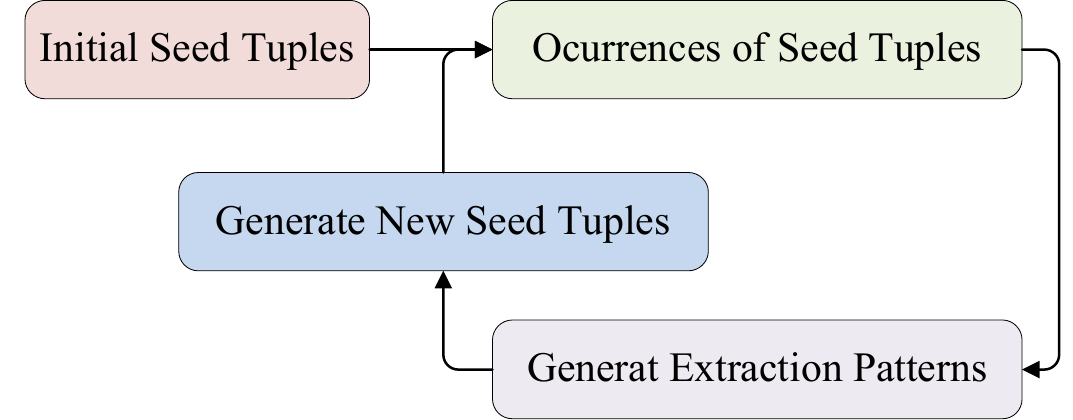}
\caption{The main idea of DIPRE.}
\label{fig1}
\end{figure*}

\textbf{Semi-supervise methods} are pattern-based methods in essence. The typical method is a bootstrapping algorithm, and the representative model is DIPRE (Dual Iterative Pattern Relation Expansion) proposed by Brin et al. \cite{brin1998extracting}. The idea behind this method is first to find some seed tuples with high confidence, the bootstrapping algorithm extracts patterns with the tuples from a large number of an unlabeled corpus. And then these patterns can be used to extract new triples. This method looks like Figure \ref{fig1}. Some other representative models are: Snowball \cite{agichtein2000snowball}, KnowItAll \cite{etzioni2004web}, TextRunner \cite{yates2007textrunner}. There is also a recent method proposed by Phi et al. \cite{phi2018ranking}.

\textbf{Unsupervised methods} adopt a bottom-up information extraction strategy based on the assumption: the context information of different entity pairs with the same semantic relation is relatively similar.
An earlier unsupervised approach is proposed by Hasegawa et al. \cite{hasegawa2004discovering}. This extraction process can be divided into three steps: extracts an entity pair and its context, clusters the entity pair according to the context, and annotates the semantic relation of each class or describes the relation type.

\textbf{Supervised methods} consider RE as a multi-class classification problem. These approaches are classified into two types: feature-based and kernel-based \cite{pawar2017relation}. In the feature-based methods \cite{rink2010utd,kambhatla2004combining}, each relation instance in the labeled data is used to train a classifier fed with subsequent new instances for classification.
Generally, these features come from useful information (including lexical, syntactical, semantic) extracted from an instance context. Without a proper feature selection, a feature-based method is difficult to improve the performance. Compared with the feature-based methods, the kernel-based \cite{bunescu2005shortest,mooney2006subsequence} methods need rarely explicit linguistic preprocessing steps. But it depends more on the performance of the kernel function designed. The key step in this approach becomes how to design an effective kernel.

\begin{figure*}
\centering
\includegraphics[width=4.1in]{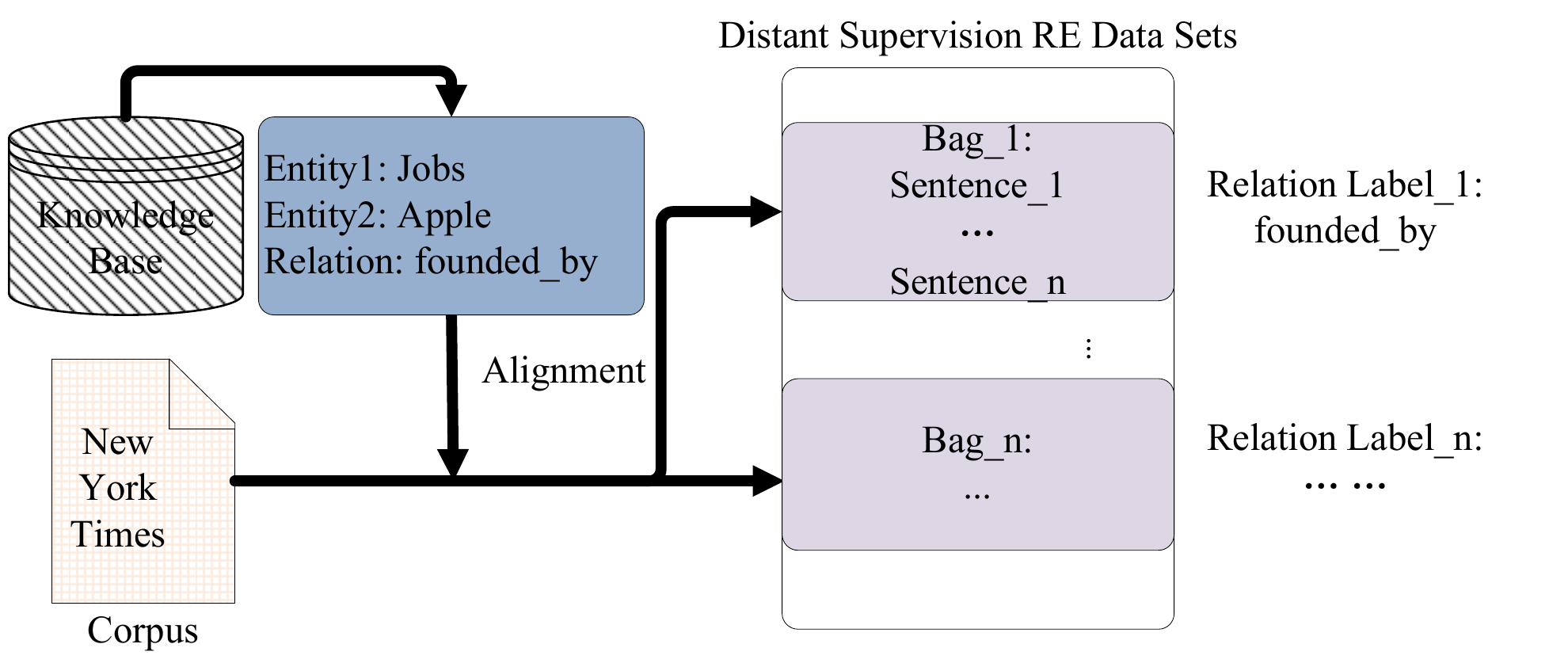}
\caption{The process of distant supervision. The upper left of Figure is the knowledge base, and the lower left is the corpus source. After the text alignment process in the middle, the right side produces the packages corresponding to the knowledge base to represent various relationships, each package represents a relationship label and these packages contain several sentence instances.}
\label{fig_ds}
\end{figure*}

\textbf{Distant Supervision methods} are a kind of knowledge-based or weakly supervised method proposed by Mintz et al. \cite{mintz2009distant}. All of these works are based on this assumption: If two entities participate in a relation, all sentences that mention these two entities can express that relation. In other words, any sentence that contains a pair of entities participating in a KB is likely to express that relation. In this way, distant supervision attempts to extract the relations between entities from the text by using a KB, such as Freebase, as the supervision source.
When a sentence and a KB refer to the same entity pair, this method marks the sentence heuristically with the corresponding relation in the KB. For example, "Jobs is the founder of Apple.", in this sentence, the person "Jobs" and the organization "Apple" appear in Freebase, and Freebase has a triple (entity1: Jobs, entity2: Apple, relation: \emph{founded\_by} ) corresponding to the mentioned entity pair. Therefore these entities express a relation \emph{founded\_by}. The process is shown in Figure \ref{fig_ds}.

\subsection{Discussion}

Although there are numerous ways to solve the problem of RE, this field has consistently shown that these methods exit various obstacles.
The hand-built pattern and semi-supervised methods require the manual exhaustion of all relation patterns, result in inevitable human errors. In the supervised method, a variety of mature NLP toolkits \cite{manning2014stanford} provide technical support for these approaches, but both feature design and kernel design are still time-consuming and laborious. The clustering results generated by unsupervised methods are generally broad, and one of the main obstacles is to define an appropriate relation inventory. Moreover, this method has limited processing capacity for a low-frequency entity pair and also lacks a standard evaluation corpus or even unified evaluation criteria. The distant supervision can effectively label data for RE, yet suffers from the wrong label and low accuracy problem. In addition, all of these approaches have domain limitations, error propagation, and poor ability to learning underlying features.

To solve these problems, some scholars try to adopt DNN-based methods to improve the performance of RE. In fact, in other fields of NLP, DNNs have been widely applied, such as machine translation, sentiment analysis, automatic summarization, question answering, and information recommendation, and all of them have achieved state-of-the-art performance. To date, DNN-based RE methods have been used in supervised and distant supervision RE mentioned above. These DNN-based methods \cite{liu2013convolution,zeng2014relation,xu2015classifying} can automatically learn features instead of manually designed features based on the various NLP toolkits. At the same time, most of them have completely surpassed the traditional methods in effect. Table \ref{tab:traditional methods} shows some comparison of traditional RE methods and earlier DNN-based methods, which illustrates that DNN-based methods can obtain higher scores with fewer features.

Based on the five traditional methods mentioned above, DNN-based methods introduced in this paper mainly focus on supervised methods and distant supervision methods.

\label{Traditional method}
\begin{center}
\begin{threeparttable}
\centering
\caption{The comparison of traditional methods and DNN-based methods.}
\label{tab:traditional methods}
\begin{tabular}{cll}
\toprule
    \textbf{Classifier}     &\textbf{Feature Sets}                &\textbf{F1}    \\
\midrule			  		
    SVM                     &\begin{tabular}[c]{@{}l@{}}POS, \\stemming,\\ syntactic patterns\end{tabular}&60.1               \\
    \hline
    SVM                     &\begin{tabular}[c]{@{}l@{}}word pair,\\ words in between\end{tabular}        &72.5               \\
    \hline
    SVM                     &\begin{tabular}[c]{@{}l@{}}POS, stemming, \\Syntactic patterns\end{tabular}  &74.8               \\
    \hline
    MaxEnt                  &\begin{tabular}[c]{@{}l@{}}POS, \\morphological,\\ noum compound, \\thesauri, \\Google n-grams, WordNet\end{tabular}
                                                                                                          &77.6               \\
    \hline
    SVM                     &\begin{tabular}[c]{@{}l@{}}POS, prefixes, \\morphological, \\WordNet, Dependency parse, \\Levin Classed, ProBank, \\FrameNet, NomLex-Plus, \\Google n-gram, \\paraphrases, TextRunner\end{tabular}                                  &82.2               \\
    \hline
    RNN                     &POS, NER ,WordNet                                                            &77.6               \\
    \hline
    MVRNN                   &POS, NER ,WordNet                                                            &82.4               \\
    \hline
    CNN+softmax             &\begin{tabular}[c]{@{}l@{}}Word pair, \\Words around word pair, \\WordNet\end{tabular}
                                                                                                          &82.7                \\
\bottomrule

\end{tabular}

    \begin{tablenotes}
      \item Some traditional classifiers, their feature sets and the F1-score for RE \cite{zeng2014relation}.
    \end{tablenotes}

\end{threeparttable}
\end{center}

\section{DNN-based Supervised RE}

\begin{figure}[htbp]
\centering
\includegraphics[width=1.0\textwidth]{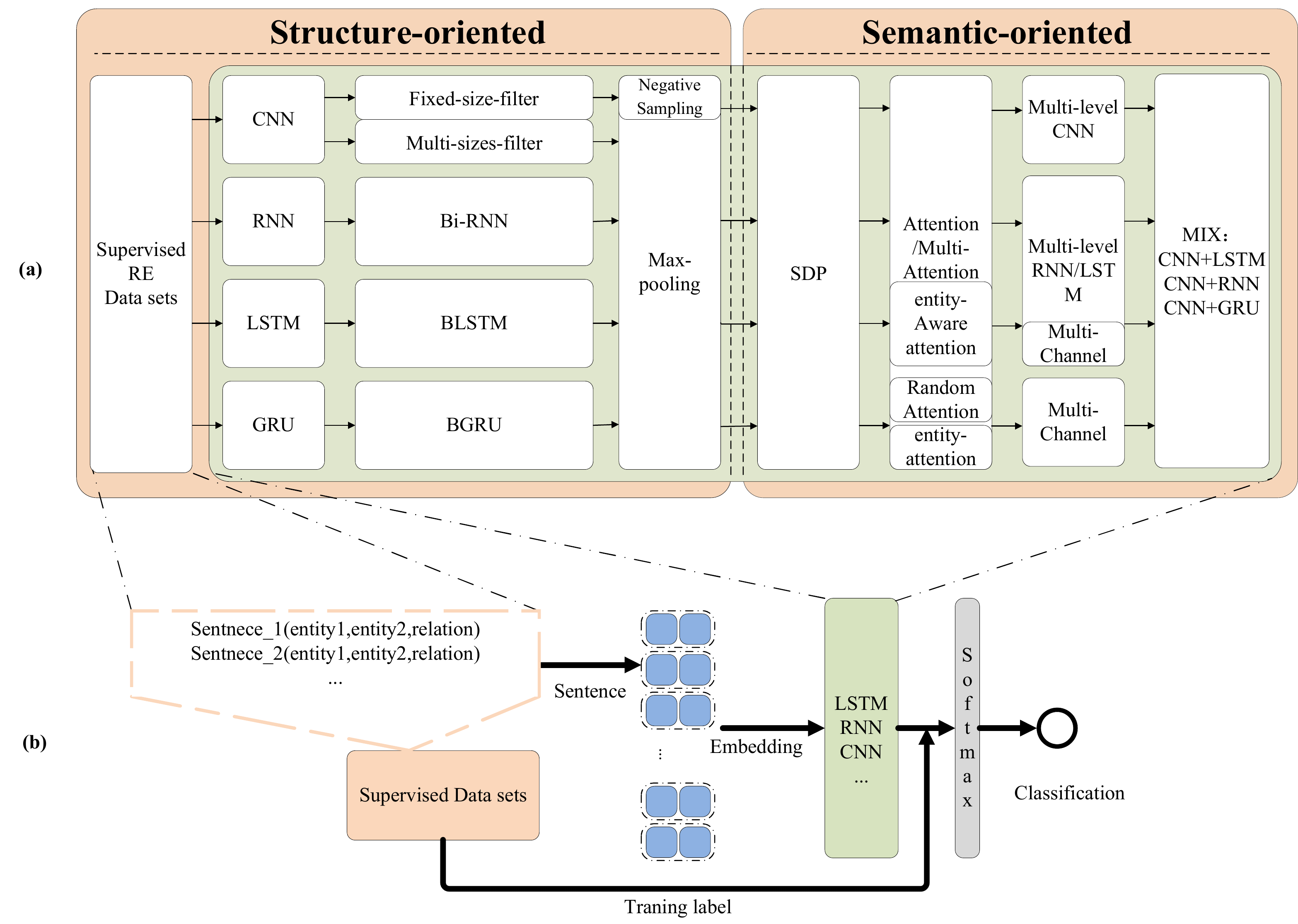}
\caption{(a) and (b) are the evolutionary process and architecture of supervised methods respectively.}
\label{envolution_of_supervised}
\end{figure}


For the sake of simplicity and clarity, this paper further subdivides the methods using different DNNs as four types (e.g., \textbf{CNN, RNN or LSTM, Mix-structure}), of which the evolutionary process is shown in Figure \ref{envolution_of_supervised} (a).
Although further subdivision is carried out, the advantages of high accuracy of DNN-based supervised methods are consistent with the traditional supervised methods, and they have been further improved.
The architecture of the supervised methods is shown in Figure \ref{envolution_of_supervised} (b). To facilitate the analysis, the development process of each type of model set is divided into two sub-types according to the evolution of model structure: \textbf{structure-oriented} and \textbf{semantic-oriented}. The structure-oriented classes improve the ability of feature extraction by changing the structure of the model; the semantic-oriented classes improve the ability of semantic representation by excavating the internal association of text. The architecture of the supervised methods is shown in Figure \ref{envolution_of_supervised} (b).


\subsection{CNN-based Methods}
CNN-based models are a general model of RE, which have achieved excellent results. In these models, a part of them play a key role in the following RE tasks, especially some modules, such as CNN with multi filters \cite{nguyen2015relation}, piecewise convolution \cite{kim2014convolutional}, attention mechanism, PE \cite{zeng2014relation}. In the following, this subsection will introduce these related models. Some comparisons are shown in Table \ref{tab:table_cnns}.

\textbf{Structure-oriented:} The first model using CNN on RE is proposed by Liu et al. \cite{liu2013convolution}. With the synonym dictionary and any other lexical features, this model transforms the sentence into a series of word vectors, which is fed to a CNN and a softmax output layer to get a classification probability. This paper is just an attempt to adopt the CNN in this task. Better result as it has achieved, this method still depends on NLP toolkits, which barely considers the semantics, model structure, and feature selection.


For improving the feature selection, Kim \cite{kim2014convolutional} introduces multiple filters and max-pooling modules for CNN, which is probably one of the earliest works in the text classification task. Based on this work, Kalchbrenneret et al. \cite{kalchbrenner2014convolutional} describe a dynamic CNN that uses a dynamic k max-pooling operator to pick up some features from the result of CNN. Both of the above two models \cite{liu2013convolution,kalchbrenner2014convolutional} achieve high performance on different tasks.

To finish this work with fewer NLP toolkits, Zeng et al. \cite{zeng2014relation} creatively put forward the concept of PE between entities, and combine the lexical information of entities with the sentence-level features extracted by the CNN, and integrate these features into the network, achieving state-of-the-art on the task of SemEval-2010task8 at 2014. This model solves the cumbersome preprocessing problem in the task and avoids the error propagation problem to some extent. After this model, almost all CNN models use the PE method, and try to extract information with fewer NLP toolkits, or even try to extract relations without using any information other than word embedding. However, with a fixed-size-filter-CNN, this model only focuses on the local features and ignores the global features.

\begin{table*}
   \resizebox{\textwidth}{!}{
   \newcommand{\tabincell}[2]{\begin{tabular}{@{}#1@{}}#2\end{tabular}}
   \begin{threeparttable}
        \centering
        \caption{The comparison of CNN-based methods.}
        \begin{tabular}{clclclcl}
            \toprule
                \textbf{Class}        & \textbf{Author}     & \textbf{Model name}       & \textbf{Framework}        & \textbf{Features set}           & \textbf{Loss function}    & \textbf{Optimization}     & \textbf{F1} \\
            \midrule
\multirow{6}[10]{*}{\textbf{Str}}   & Liu et al. \cite{liu2013convolution}  & -                 & fixed-size-filter-CNN     & One-hot                         & -                         & -                         & -          \\
                            \cmidrule{2-8}
                            & \multirow{2}[4]{*}{Zeng et al. \cite{zeng2014relation}} & -  & \multirow{2}[4]{*}{fixed-size-filter-CNN} & WE1+WA+WN+PE & \multirow{2}[4]{*}{Cross entropy} & \multirow{2}[4]{*}{SGD} & 82.7 \\
                            \cmidrule{3-3} \cmidrule{5-5} \cmidrule{8-8}
                            &                                    & -                         &                            & WE1+WA+WN                      &                           &                           & 69.7 \\
                            \cmidrule{2-8}
                            & Nguyen et al. \cite{nguyen2015relation} & -               & \tabincell{l}{multi-sizes-filter-CNN\\+max-pooling}   & WE2+PE  & -                         & SGD                       & 82.8 \\
                            \cmidrule{2-8}
                            &\multirow{2}[4]{*}{Santos et al. \cite{santos2015classifying}}&CR-CNN&\multirow{2}[4]{*}{\tabincell{l}{fixed-size-filter-CNN\\+max-pooling}}&\multirow{2}[4]{*}{WE2+PE}&Ranking loss& \multirow{2}[4]{*}{SGD}& 84.1 \\
                            \cmidrule{3-3}\cmidrule{6-6}\cmidrule{8-8}
                            &               & -                  &                           &                                                             & Cross entropy             &                           & 82.4 \\
                            \cmidrule{1-8}
\multirow{4}[20]{*}{\textbf{Sem}}    & \multirow{2}[8]{*}{Xu et al. \cite{xu2015semantic}} & depLCNN+NS & \tabincell{l}{fixed-size-filter-CNN\\+max-pooling\\+Negative Sampling} & WE1+WA+WN+SDP & \multirow{2}[4]{*}{Cross entropy} & \multirow{2}[4]{*}{SGD} & 85.6 \\
                            \cmidrule{3-5}\cmidrule{8-8}
                            &               & -                     & \tabincell{l}{fixed-size-filter-CNN\\+max-pooling}  & WE1+PE                         &                           &                           & 83.7 \\
                            \cmidrule{2-8}
                            & \multirow{2}[12]{*}{Wang et al. \cite{wang2016relation}} & -& \tabincell{l}{fixed-size-filter-CNN\\+two-level-CNN\\+two-level-attention\\+max-pooling} & WE2+PE+WA & \multirow{2}[12]{*}{Distance function} & \multirow{2}[12]{*}{SGD} & 88 \\
                            \cmidrule{3-5}\cmidrule{8-8}
                            &       & Att-Pooling-CNN & \tabincell{l}{fixed-size-filter-CNN\\+two-level-CNN\\+max-pooling} & WE2+PE+WA                     &                           &                           & 86.1 \\
            \bottomrule
         \end{tabular}
         \begin{tablenotes}
              \item In this tabel, the Str, Sem refer to structure-oriented classes, and semantic-oriented classes. The WE1, WE2, WE3 refer to Word embedding probosed by Turian et al. \cite{turian2010word}, Mikolov et al. \cite{mikolov2013efficient}, Pennington et al. \cite{pennington2014glove} respectively. The WA, WN, PE, PI refer to Word around nominals, WordNet, Position embedding, Position indicator. The SDP, GR, WNSYN, Relative-DEP refer to shortest dependency path, grammar relation, hypernyms, relative-dependency. The F1 value are based on semeval2010-task8 \cite{hendrickx2009semeval}. This table lists the best result of the model and its variants, and subsequent tables follow this format.
         \end{tablenotes}
         \label{tab:table_cnns}
   \end{threeparttable}
   }
   \end{table*}

\begin{figure*}[htbp]
\centering
\includegraphics[angle=-0,width=1.0\textwidth]{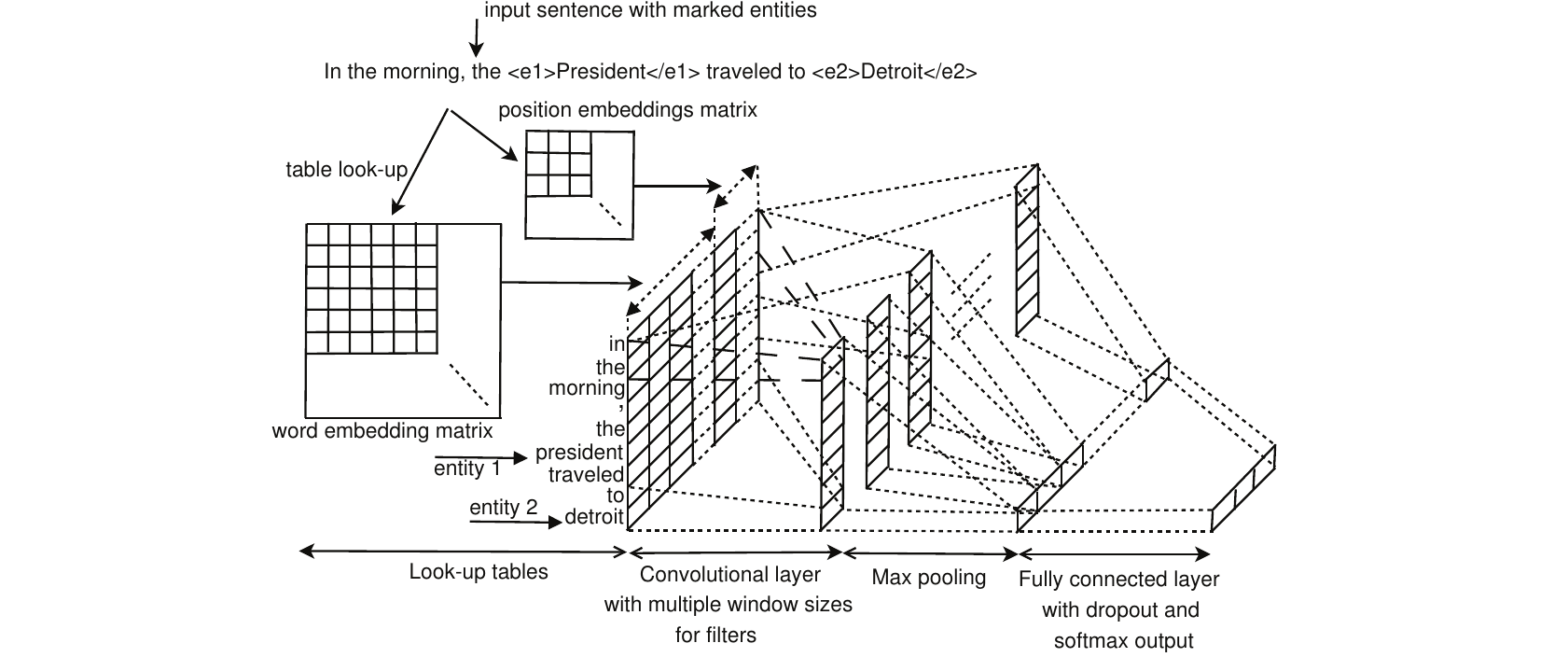}
\caption{The structure of model \cite{nguyen2015relation} with multi window filters.}
\label{nguyen2015relation}
\end{figure*}

To bring more structure information, Nguyen et al. \cite{nguyen2015relation} combine the concept of multiple-window size filters for RE based on Kim's \cite{kim2014convolutional} work. Compared with fixed-size-filter-CNN, this paper demonstrates that multi window sizes filters (multi-sizes-filter-CNN) bring more structured information to the model. This would be an efficient way to improve the CNN architecture. Many subsequent models will adopt this technique as well. Meanwhile, this paper also shows that the word vectors pre-trained and changing dynamically with model training are helpful to improve the performance. But in fact, dynamic word vector is not often referenced. The model structure is shown in Figure \ref{nguyen2015relation}.
Besides, all the above models deploy the softmax module, which cannot eliminate the influence of other similar classes affecting the final classification results.

Santos et al. \cite{santos2015classifying} improve the loss function instead of the softmax classifier. The Model's parameters are trained by minimizing a new ranking loss function (CR-CNN) over the training set, giving a higher probability to the correct class and lower probability for the wrong classes. This new loss function improves this model and could be used in other classifiers.


\textbf{Semantic-oriented:} To learn more robust relation representations, Xu et al. \cite{xu2015semantic} propose a model through a CNN with the SDP (talked about it in Sec \ref{the shortest dependency path}). The model, which mostly takes the subject to object of a sentence as input, remove the words that are not related to the relation discrimination, following high accuracy with simply negative samples. This is the first DNNs model using the SDP, and this technic and its variants will be widely adopted in subsequent models.

Another way to improve the semantic representation is attention mechanism. Wang et al. \cite{wang2016relation} use two levels of attention mechanism, called Multi-Level Attention CNNs which enables end to end learning from task-specific labeled data. This multiple attention mechanism considers both the semantic information at the word level and sentence level. This model rarely uses any other external semantic information at all. At the first level attention, the model constructs an entity-based attention matrix at the input level, which labels the words related to the corresponding relation. At the second level, the mechanism captures more abstract high-level features to construct the final output matrix.



\subsection{RNN or LSTM based Methods}

\begin{table*}
 \resizebox{\textwidth}{!}{
    \newcommand{\tabincell}[2]{\begin{tabular}{@{}#1@{}}#2\end{tabular}}
    \begin{threeparttable}
        \centering
        \caption{The comparison of RNN-based methods.}
        \begin{tabular}{cllp{13em}llll}
        \toprule
            \multicolumn{1}{l}{\textbf{Class}} & \textbf{Author} & \textbf{Model name} & \multicolumn{1}{l}{\textbf{Framework}} & \textbf{Features set} & \textbf{Loss function} & \textbf{Optimization} & \textbf{F1} \\
        \midrule
            \multicolumn{1}{c}{\multirow{4}[12]{*}{\textbf{Str}}} & \multirow{2}[4]{*}{Zhang et al. \cite{zhang2015relation}} & \multirow{2}[4]{*}{RNN} & \multicolumn{1}{l}{\multirow{2}[4]{*}{\tabincell{l}{Bi-RNN\\+max-pooling}}} & WE1+PI & \multirow{2}[4]{*}{Cross entropy} & \multirow{2}[4]{*}{SGD} & 80 \\
            \cmidrule{5-5} \cmidrule{8-8} &       &       &       & WE2+PI &       &       & 82.5 \\
            \cmidrule{2-8}
            & \multirow{2}[4]{*}{Zhang et al. \cite{zhang2015bidirectional}} & \multirow{2}[4]{*}{BLSTM} & \multicolumn{1}{l}{\multirow{2}[4]{*}{\tabincell{l}{BLSTM\\+max-pooling}}} & \multicolumn{1}{l}{\tabincell{l}{WE3+PE+WN+NER\\+POS+WNSYN\\+Relative-DEP}} & \multirow{2}[4]{*}{-} & \multirow{2}[4]{*}{-} & 84.3 \\
            \cmidrule{5-5}  \cmidrule{8-8}        &       &       &       & WE3+WA &       &       & 82.7 \\
            \cmidrule{1-8}
            \multicolumn{1}{c}{\multirow{12}[40]{*}{\textbf{Sem}}}& \multirow{2}[4]{*}{Xu et al. \cite{xu2015classifying}} & \multicolumn{1}{l}{\multirow{2}[4]{*}{SDP-LSTM}} & \multirow{2}[4]{*}{SDP+LSTM} & WE2+SDP & \multirow{2}[4]{*}{Cross entropy} & \multirow{2}[4]{*}{SGD} & 82.4 \\
            \cmidrule{5-5}  \cmidrule{8-8}        &       &       &       & WE2+SDP+WA+POS+GR &       &       & 83.7 \\
            \cmidrule{2-8}
            & \multirow{2}[4]{*}{Zhou et al. \cite{zhou2016attention} } & Att-BLSTM & \multicolumn{1}{l}{\tabincell{l}{BLSTM\\+attention}} & WE3+PI & \multicolumn{1}{l}{\tabincell{l}{Negative-log-likelihood\\(Cross-Entropy)}} & AdaDelta & 84 \\
            \cmidrule{3-8}  \cmidrule{8-8}        &       & BLSTM & \multicolumn{1}{l}{\tabincell{l}{BLSTM\\+max-pooling}} & WE3+WA & -     & -     & 82.7 \\
            \cmidrule{2-8}
            & \multirow{2}[4]{*}{Xu et al. \cite{xu2016improved}} & \multirow{2}[4]{*}{DRNNs} & \multicolumn{1}{l}{\tabincell{l}{multi-channel-rnn\\+max-pooling\\+augmentation}} & \multicolumn{1}{l}{\tabincell{l}{WE2+GR+POS\\+WN+SDP(augmentation)}} & \multirow{2}[4]{*}{ Cross entropy} & \multirow{2}[4]{*}{SGD} & 86.1 \\
            \cmidrule{4-5}  \cmidrule{8-8}        &       &       & \multicolumn{1}{l}{\tabincell{l}{multi-channel-rnn\\+max-pooling}} & WE2+GR+POS+WN+SDP &       &       & 84.16 \\
            \cmidrule{2-8}
            & \multirow{2}[4]{*}{Xiao et al. \cite{xiao2016semantic}} & \multirow{2}[4]{*}{BLSTM+BLSTM} & \multicolumn{1}{l}{\multirow{2}[4]{*}{\tabincell{l}{2-level-BLSTM\\+attention}}} & WE2+WN+NER & \multirow{2}[4]{*}{Ranking loss function} & \multirow{2}[4]{*}{AdaGrad} & 84.27 \\
            \cmidrule{5-5}  \cmidrule{8-8}         &       &       &       & WE2   &       &       & 83.9 \\
            \cmidrule{2-8}
            & \multirow{2}[8]{*}{Qin et al. \cite{qin2017designing}} & EAtt-BiGRU & \multicolumn{1}{l}{Bi-GRU\newline{}+entity-attention} & \multicolumn{1}{l}{\multirow{2}[8]{*}{WE2+PE}} & \multirow{2}[4]{*}{-} & \multirow{2}[8]{*}{AdaDelta } & 84.7 \\
            \cmidrule{3-4} \cmidrule{8-8} &       & -     & \multicolumn{1}{l}{\tabincell{l}{Bi-GRU\\+random-att}} &       &       &       & 83.6 \\
            \cmidrule{2-8}
            & \multirow{2}[4]{*}{Zhang et al. \cite{zhang2019multi}} & BiGUR-MCNN-ATT & \multicolumn{1}{p{13em}}{BiGRU\newline{}+multi-size-filter-cnn-attention} & \multicolumn{1}{l}{\multirow{2}[4]{*}{WE2+PE+SDP}} & \multirow{2}[4]{*}{-} & \multirow{2}[4]{*}{AdaDelta} & 84.7 \\
            \cmidrule{3-4}   \cmidrule{8-8}        &       & BiGUR-Random-ATT & \multicolumn{1}{l}{\tabincell{l}{BiGRU\\+random-attention}} &       &       &       & 84.2 \\
            \cmidrule{2-8}
            & \multicolumn{1}{c}{\multirow{2}[4]{*}{Lee et al. \cite{lee2019semantic}}} & BLSTM+LET & \multicolumn{1}{l}{\tabincell{l}{BLSTM\\+Entity-aware-attention\\+Latent-Entity-Type}} & \multicolumn{1}{l}{WE2+PE+LET} & \multirow{2}[4]{*}{Cross entropy} & \multirow{2}[4]{*}{AdaDelta} & 85.2 \\
            \cmidrule{3-5}   \cmidrule{8-8}        &       & -     & \multicolumn{1}{l}{\tabincell{l}{BLSTM\\+Entity-aware-attention}} & \multicolumn{1}{l}{WE2+PE} &       &       & 84.7 \\
        \bottomrule
        \end{tabular}%
        \begin{tablenotes}
            \item Table symbols have the same meanings as Table. \ref{tab:table_cnns}.
        \end{tablenotes}
        \label{tab:tabel_rnns}%
    \end{threeparttable}
    }
 \end{table*}

The general problems of CNNs in RE are that CNNs rarely consider global features and time sequence information, especially for the long-distance dependency between entity pair. RNNs, LSMTs, and GRUs, the latest research methods for sequence modeling and problem transformation can alleviate these problems. In this section, this paper will introduce some RNN-based methods, of which the comparisons are shown in Table \ref{tab:tabel_rnns}.


\textbf{Structure-oriented:} For learning relations within a long context and considering the timing information, Zhang et al. \cite{zhang2015relation} use a bi-directional RNN architecture to this task. RNN combines the output of each hidden state and then represents the feature at the sentence level. At the end of the model, it conducts a max-pooling operation to pick up a few trigger word features for prediction. Although max-pooling operation simplifies feature extraction, the effectiveness of these features remains to be discussed. Besides, the RNN model still has the problem of gradient explosion. To solve the problem of the gradient explosion, LSTM \cite{hochreiter1997long} is proposed by using the gate mechanism.
Based on this, Xu et al. \cite{xu2015classifying} propose a model with LSTM (will be discussed in the semantic-oriented type).

With complete, sequential information about all words in the sentence is beneficial to RE, Zhang et al. \cite{zhang2015bidirectional} apply the bi-directional long short-term memory networks (BLSTM) to obtain the sentence level representation, and also use several lexical features. The experiment results show that using word embedding as input features alone is enough to achieve state-of-the-art results. This study documents the effectiveness of the BLSTM. Although the method improves the representation of sentence-level features, there are still two problems: a large number of external artificial features are introduced, and no effective feature filter mechanism.

\begin{figure*}[htbp]
\includegraphics[width=1.0\textwidth]{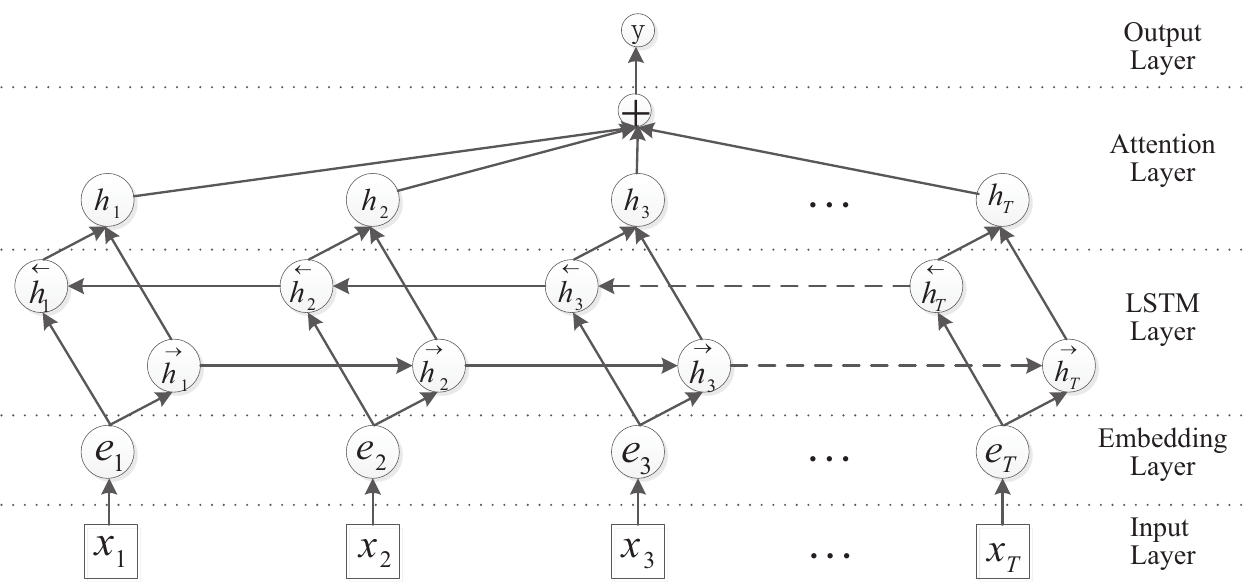}
\caption{The structure of the model \cite{zhou2016attention}, BLSTM with an attention mechanism.}\label{zhou2016attention}
\label{zhou2016attention}
\end{figure*}

\textbf{Semantic-oriented:} Based on the above problems, Xu et al. \cite{xu2015classifying} propose a new DNNs model called SDP-LSTM.
This model leverages four types of information: Word vectors, POS tags, Grammatical relations, and WordNet hypernyms, to construct four channels for this model to support external information.  And then, it concatenates the result of the four channels to the softmax layer for prediction.
This model is a little more complex than Zhang et al. \cite{zhang2015relation} by considering a lot of additional syntaxes and semantic information.

Follow the above work SDP-LSTM \cite{xu2015classifying}, to overcome the problem of shallow architecture that hardly represent the potential space in different network levels, Xu et al. \cite{xu2016improved} increase the neural network layers to tackle this challenge, with which this model captures the abstract features along the two sub-paths of SDP. Meanwhile, the small size of the semeval2010-task8 set and deeper neural networks may easily result in overfitting. Hence, the author augments the data set by adding the directivity of the data based on the original SDP, avoiding the overfitting problem.

The SDP filters the input text but can not filter the extracted features. To tackle this issue, Zhou et al. \cite{zhou2016attention} come up with the attention mechanism in BLSTM, which automatically highlight the important features only with the raw text instead of any other NLP toolkits or lexical resources. This work is a representative BLSTM model and the architecture is shown in Figure \ref{zhou2016attention}. Similar to the work of Zhou et al. \cite{zhou2016attention}, Xiao et al. \cite{xiao2016semantic} propose a two-level BLSTM architecture with a two-level attention mechanism to extract a high-level representation of the raw sentence.


Although the attention mechanism gives more weight to the important features extracted by the model, this work \cite{zhou2016attention}  just presents a random weight, which lacks the consideration of the prior knowledge. Therefore, the following works improve this model.

Fed with entity pair and sentence, EAtt-BiGRU proposed by Qin et al. \cite{qin2017designing} leverage the entity pair as prior knowledge to form attention weight. Different from Zhou et al.'s \cite{zhou2016attention} work, EAtt-BiGRU applies bi-directional GRU (BiGRU) instead of BLSTM to reduce computation, which helps in obtaining the representation of sentence and adopts a one-way GRU to extract prior knowledge of entity pair. With the representation and prior knowledge, this model can generate the corresponding attention weight adaptively. This work improves the random attention mechanism, but how to better integrate the prior knowledge needs further studying.

Zhang et al. \cite{zhang2019multi} propose another kind of attention mechanism based on the SDP, which is another prior knowledge. This model uses Bi-GRU to extract sentence-level features and attention weights to select features from multi-channel CNN for classifying. Compared with other random or entity based attention mechanisms \cite{zhou2016attention,qin2017designing}, this model constructs a better attention weight using the SDP.

Further research is followed by Lee et al. \cite{lee2019semantic} who propose a mixed model with BLSTM, self-attention, entity-aware attention, and latent entity typing module, getting state-of-the-art without any high-level features.
In general, the instance of the data set has no attribute of the entity type. However, the entity pair type is closely related to the relation classes. Previous works can only get word-level or sentence-level attention, but rarely obtain the degree of correlation between entities and other related words. Hence, this model introduces the latent entity typing module, self-attention module, entity-aware attention module, giving more prior knowledge.

\subsection{Mix-structure based Methods}

In addition to the above two types of models, some scholars combine these models based on their respective characteristics, which can be beneficial to RE task.
There also exist two ways to merge these models: simply combination (\textbf{Structure-oriented}) \cite{zhang2018relation} and attention mechanism (\textbf{Semantic-oriented}) \cite{guo2019single}. The comparison of them are shown in Table \ref{tab:table_mix}.

\begin{table*}
\resizebox{\textwidth}{!}{
\newcommand{\tabincell}[2]{\begin{tabular}{@{}#1@{}}#2\end{tabular}}
\begin{threeparttable}
\centering
\caption{The comparison of methods based on Mix neural networks.}
\begin{tabular}{clclclcl}
\toprule
    \textbf{Class} & \textbf{Author} & \textbf{Model name} & \textbf{Framework} & \textbf{Features set} & \textbf{Loss function} & \textbf{Optimization} & \textbf{F1} \\
\midrule
\multirow{5}[11]{*}{\textbf{Str}}
                  & \multirow{2}[4]{*}{Zheng et al. \cite{zheng2016neural}}&MixCNN+CNN&\tabincell{l}{multi-sizes-filter-CNN\\+fixed-size-filter-CNN\\+max-pooling}&\multirow{2}[8]{*}{WE2+WA}&\multirow{2}[8]{*}{Cross entropy} & \multirow{2}[8]{*}{SGD} & 84.8 \\
                  \cmidrule{3-4}\cmidrule{8-8}
                  &                       & MixCNN+LSTM & \tabincell{l}{multi-sizes-filter-CNN\\+LSTM\\+max-pooling} &           &                        &                       & 83.8 \\
                  \cmidrule{2-8}
                  & \multirow{3}[6]{*}{Zhang et al. \cite{zhang2018relation}} & BLSTM-CNN & \multirow{3}[6]{*}{\tabincell{l}{BLSTM\\+fixed-size-filter-CNN\\+max-pooling}}& WE2+PI+PE & \multirow{3}[6]{*}{-} & \multirow{3}[6]{*}{-} & 83.2 \\
                  \cmidrule{3-3}\cmidrule{5-5}\cmidrule{8-8}
                  &                       & BLSTM-CNN+PF                           &                    & WE2+PE                &                        &                       & 81.9 \\
                  \cmidrule{3-3}\cmidrule{5-5}\cmidrule{8-8}
                  &                       & BLSTM-CNN+PI                           &                    & WE2+PI                &                        &                       & 82.1 \\
                  \cmidrule{1-8}
\multirow{7}[40]{*}{\textbf{Sem}}
                   & \multirow{2}[4]{*}{Cai et al. \cite{cai2016bidirectional}}& \multirow{2}[4]{*}{BRCNN}&\multirow{2}[4]{*}{\tabincell{l}{2level-LSTM\\+2level-fixed-size-filter-CNN\\+SDP(2direction)}} & WE2+POS+NER+SDP & \multirow{2}[4]{*}{Cross entropy} & \multirow{2}[4]{*}{AdaDelta} & 86.3 \\
                  \cmidrule{5-5}\cmidrule{8-8}
  \rule{0pt}{12pt}&                       &                      &                                       & WE2+SDP               &                        &                       & 85.4 \\
                  \cmidrule{2-8}
                  & \multirow{2}[4]{*}{Ren et al. \cite{ren2018neural}}&DesRC(BRCNN)&\tabincell{l}{BRCNN\\+fixed-size-filiter-CNN-description\\+attention}&We2+PE+WN+SDP&\multirow{2}[4]{*}{-}&\multirow{2}[4]{*}{SGD} & 87.4 \\
                  \cmidrule{3-5}\cmidrule{8-8}
                  &                       & -                    & \tabincell{l}{BRCNN\\+fixed-size-filiter-CNN} & We2+PE+SDP    &                        &                       & 84.7 \\
                  \cmidrule{2-8}
                  & \multirow{2}[4]{*}{Guo et al. \cite{guo2019single}} & Att-RCNN & \tabincell{l}{fixed-size-filter-CNN \\ +max-pooling \\ +BiGRU \\ +attention } & \multirow{2}[4]{*}{WE2+SDP} & \multirow{2}[4]{*}{Pairwise logistic loss } & \multirow{2}[4]{*}{SGD} & 86.6 \\
                  \cmidrule{3-4}\cmidrule{8-8}
                  &                       & -                   & \tabincell{l}{fixed-size-filter-CNN\\+max-pooling\\+BiGRU}&   &                        &                       & 85.1 \\
                  \cmidrule{2-8}
                  & \multirow{1}[1]{*}{Wang et al. \cite{wang2020direction}} & Bi-SDP & \tabincell{l}{fixed-size-filter-CNN \\ +max-pooling \\ +BLSTM \\ +attention } & \multirow{1}[1]{*}{WE2+PE+SDP} & \multirow{1}[1]{*}{Cross entropy} & \multirow{1}[1]{*}{SGD} & 85.1 \\

\bottomrule
\end{tabular}%
\begin{tablenotes}
    \item Table symbols have the same meanings as Table \ref{tab:table_cnns}.
\end{tablenotes}
\label{tab:table_mix}%
\end{threeparttable}%
}
\end{table*}

\textbf{Structure-oriented:} To integrate RNN and CNN, Zheng et al. \cite{zheng2016neural} propose two neural networks based on CNN and LSTM (MixCNN+CNN and MixCNN+LSTM) framework by joint learning the entity semantic and relation pattern. In this model, the entity semantic properties can be reflected by their surrounding words, which can tackle the unknown words in entities (Out-of-vocabulary problem), and the relation pattern modeled by the sub-sentence between the given entities instead of the whole sentence. With the entity semantic and relation pattern, the performance of RE can be improved.
This research sheds new light on merging these two modules, showing the complementarity of the two modules as well as the necessity of module integration.

Zhang et al. \cite{zhang2018relation} introduce the BLSTM-CNN, without any lexical attention mechanism or NLP toolkits, just utilize three kinds of resources, word embedding, PE, and PI, showing that simply merging BLSTM and CNNs can perform better than any other single models. However, the PE and PI, showing the words around the nominals, are same functions for RE which may result in overlap feeding.

\textbf{Semantic-oriented:} Different from the above two works, Cai et al. \cite{cai2016bidirectional} combine these two types of models depending on the SDP. To improve the model's sense of relation directivity, this model, called BRCNN, learn sentence features from the SDP on both positive and negative directions, which is beneficial for predicting the direction of the relation.

Following Cai et al.'s \cite{cai2016bidirectional} work, Ren et al. \cite{ren2018neural} advance a further work, a traditional CNN architecture with BRCNN \cite{cai2016bidirectional}, using two kinds of attention ('intra-cross') to combine the classification features that come from the original sentences and their corresponding descriptions. The description of an entity is from the external text, which can enrich the prior knowledge. Compared with different experiments and models, the result demonstrates that text descriptions can provide more features to model and replace WordNet to a certain extent in RE. In this line, this is the first RE method with entity description information. But the method of extracting description is too simple. As an external knowledge, description information should be closely related to the original sentence, therefore the selection of description should be more targeted.

Guo et al. \cite{guo2019single} propose a novel Att-RCNN model to extract text features. This model leverages GRU units instead of LSTM units, which has a higher speed in computing convergence and a more efficient CNNs to extract high-level features. Meanwhile, the two-level attention mechanism is similar to \cite{wang2016relation}. The special part of this model is the introduction of a new de-noise method, which can get a continuous fragment of the original text, based on the SDP \cite{xu2015classifying}.

Wang et al. \cite{wang2020direction} further utilize the SDP to construct a Bi-SDP with a parallel attention weight to cope the direction of relation. In a sense, this method introduces more information about the SDP, and interprets another function (giving a hint of the direction of one relation) of a preposition in a sentence, which is ignored by previous works.

\subsection{Discussion}
As stated above, the structure-oriented and semantic-oriented models improve the RE task in different aspects: one improves the ability of feature extraction, and the another focus on the semantic representation of text. From the Table \ref{tab:table_cnns}, \ref{tab:tabel_rnns}, and \ref{tab:table_mix}, they show that semantic-oriented models exhibit more effective than structure-oriented models. This phenomenon has been widely observed in all of these models, which reflects that relational facts in a sentence are semantically strongly related to the sentence itself. In other words, we argue that the constructed models around SDP, providing external semantic information, give a interpretable way of human-level thinking and cognition to cope the RE task. Except the above works, some other research filed also give new light on RE, such as: distilling knowledge \cite{zhang2020distilling} and auxiliary learning \cite{lyu2020auxiliary}. Utilizing distilling knowledge method to generate soft labels and guide the student network to learn dark knowledge, which seems can overcome the limitation of relational inventory or hard-label problem in supervised method. Auxiliary learning provides a simple module to further dig out the latent semantic relation in the wrong classified result, which can alleviate the semantic gap between the sentence and the label. All of the above methods provide a solid foundation for the development of supervised RE. However, we have to admit that there are still many limitations in these approaches, which would be mitigated by distant supervision.



\section{DNN-based Distant Supervision RE}

\begin{table*}
\resizebox{\textwidth}{!}{
\newcommand{\tabincell}[2]{\begin{tabular}{@{}#1@{}}#2\end{tabular}}
\begin{threeparttable}
  \centering
  \caption{The comparison of DNN-based distant supervision methods.}
    \begin{tabular}{clclcccl}
    \toprule
        \textbf{Class} & \textbf{Author} & \textbf{Model name} & \textbf{Framework} & \textbf{Features set} & \textbf{De-noise method} & \textbf{Loss function} & \textbf{Optimization} \\
    \midrule
\multirow{2}[8]{*}{\textbf{En}}& Zeng et al. \cite{zeng2015distant}  & PCNN  & \tabincell{l}{multi-fixed-size-filter-CNN\\+piecewise-max-pooling} & WE2+PE   & Multi-instance Learning & Cross entropy & Adadelta \\
        \cmidrule{2-8}
             & Jiang et al. \cite{jiang2016relation} & MIMLCNN & \tabincell{l}{multi-fixed-size-filter-CNN\\+piecewise-max-pooling\\+cross-sentence-max-pooling} & WE2+PE   & Output multi label & Cross entropy & Adadelta \\
        \cmidrule{1-8}
\multirow{5}[30]{*}{\textbf{Re}}& Yang et al. \cite{yang2017multi} & BiGRU+2ATT & \tabincell{l}{BiGRU\\+word-level-attention\\+sentence-level-attention} & WE2+PE   & Sentence-level-attention & Cross entropy & Adam \\
        \cmidrule{2-8}
             & Lin et al. \cite{lin2017neural} & MNRE  & \tabincell{l}{multi-fixed-size-filter-CNN\\+max-pooling\\+mono-lingual attention\\+cross-lingual attention} & WE1+PE & \tabincell{l}{Mono-lingual\\(Sentence-level- attention)} & -     & SGD \\
        \cmidrule{2-8}
        & Lin et al. \cite{lin2016neural} & PCNN+ATT & \tabincell{l}{multi-fixed-size-filter-CNN\\+piecewise-max-pooling\\+sentence-level-attention} & WE2+PE   & Selective attention & Cross entropy & SGD \\
        \cmidrule{2-8}
             & Banerjee et al. \cite{banerjee2018relation}  & MEM   & multi-channel-BLSTM & \tabincell{l}{WE3+DP+POS} & Co-occurrence statistics & Cross entropy & - \\
        \cmidrule{2-8}
             & Du et al. \cite{du2018multi} & MLSSA & \tabincell{l}{BLSTM\\+word-level-attention\\+sentence-level-attention} & WE2+PE   & Sentence-level-attention & -     & Adam \\
        \cmidrule{1-8}
\multirow{3}[22]{*}{\textbf{Ex}}& Ji et al. \cite{ji2017distant} & APCNNs+D & \tabincell{l}{multi-fixed-size-filter-CNN\\+piecewise max-pooling\\+fixed-size-filiter-CNN-description\\+sentence-level-attention} & WE2+PE   & Sentence-level-attention & Cross entropy & Adadelta \\
        \cmidrule{2-8}
             & Wang et al. \cite{wang2018label} & LFDS  & \tabincell{l}{multi-fixed-size-filter-CNN\\+piecewise max-pooling\\+word-level-attention} & WE2+PE   & KG Embedding & Margin loss & - \\
        \cmidrule{2-8}
             & Vashishth et al. \cite{vashishth2018reside} & RESIDE & \tabincell{l}{GCN\\+BiGRU\\+word-level-attention\\+senten-level-attention} & WE3+PE+DP & -     & -     & - \\
        \cmidrule{1-8}
\multirow{2}[5]{*}{\textbf{Pl}}& Qin et al. \cite{qin2018robust} & RL    & \tabincell{l}{fixed-size-filter-CNN\\+Reinforcement learning} & WE2   & Reinforcement Learning & -     & - \\
        \cmidrule{2-8}
               & Qin et al. \cite{qin2018dsgan} & DSAGN & \tabincell{l}{fixed-size-filter-CNN\\+GAN} & WE2+PE   & DSGAN & -     & - \\
    \bottomrule
    \end{tabular}%
    \begin{tablenotes}
        \item Table symbols have the same meanings as Table \ref{tab:table_cnns}. In addition, the En, Re, Ex, Pl refer to Sentence encoder, Enhanced representation, External knowledge, Plug-and-play component.
    \end{tablenotes}
    \label{tab:table_DS}%
\end{threeparttable}%
}
\end{table*}

\begin{figure*}[htbp]
\includegraphics[width=1.0\textwidth]{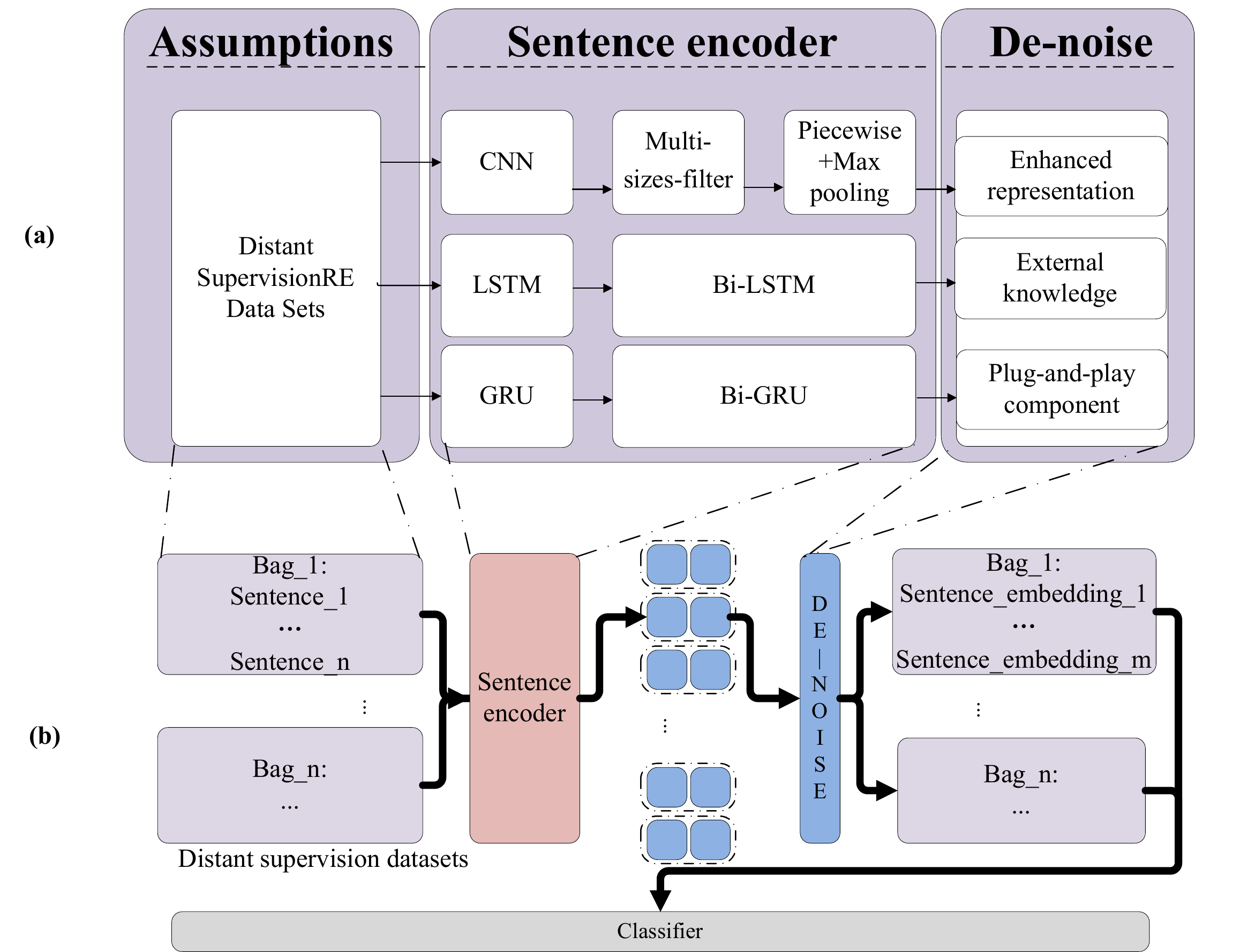}
\caption{(a) and (b) are the evolutionary process and architecture of DNN-based distant supervision methods. The sub-figure (b) also shows the relationship between sentence encoder and de-noise method. The sentence encoder is responsible for encoding the sentence bag in the distant supervision data set into vectors, while the de-noise algorithm is responsible for selecting the sentences in each bag that can correctly represent the relationship.}
\label{envolution_of_DS}
\end{figure*}

This paper introduces the conception of distant supervision in section \ref{Traditional method}.
In the supervised method of RE, the insufficient training corpus puzzles the further development of RE despite of their excellent results.
To solve this problem, Mintz et al. \cite{mintz2009distant} propose distant supervision, which is strongly based on an assumption that plays an important role in the selection of training examples. To date, this assumption has evolved into \textbf{Three assumptions}.

To solve the insufficient training samples problem, in 2009, Stanford University Professor Mintz et al. \cite{mintz2009distant} proposed RE method of distant supervision at ACL conference. This method rarely needs manual annotation and generates large-scale training data sets (talked about that in section \ref{Traditional method}). The large-scale training data sets are from these assumptions:

\textbf{Assumption 1:}
\emph{If two entities participate in a relation, all sentences that mention these two entities express that relation. }

\textbf{Assumption 2:}
\emph{If two entities participate in a relation, at least one sentence that mentions these two entities might express that relation. }

\textbf{Assumption 3:}
\emph{A relation holding between two entities can be either expressed explicitly or inferred implicitly from all sentences that mention these two entities. }

The first assumption \cite{mintz2009distant} physically aligns the texts with a KB and trains a classifier heuristically using the existing triples in the KB. Riedel et al. \cite{riedel2010modeling} thought the first assumption is too strong resulting in the wrong label problem (noisy problem), and then proposed the second assumption, called: "one sentence from one bag". It leads to more accurate results. The third assumption \cite{jiang2016relation} can consider more sentence features.

Based on the above three assumptions, distant supervision methods are composed of two research directions: \textbf{Sentence Encoder} optimizes the model and the performance of RE; \textbf{De-noise Algorithm} improves the quality of the data sets, which can be further subdivided into three main ways: the first makes full use of the word-level and sentence-level features of the instances in the bag for \textbf{enhancing representation}; the second introduces the \textbf{external knowledge}; while the third constructs a \textbf{plug-and-play component}. This section presents some related works and summarizes the evolution of this approach (shown in Figure \ref{envolution_of_DS} (a)). The whole architecture of DNN-based distant supervision is shown in Figure \ref{envolution_of_DS} (b). Hence, the rest of this chapter will follow these four points: 1) encoder-based, 2) representation-based, 3) knowledge-based, 4) plug-and-play based. In addiction, different from the supervised method, Table \ref{tab:table_DS} does not include evaluation indicators. Because most of these methods in Table \ref{tab:table_DS} is just an approximate measure of precision.

\subsection{Encoder-based Methods}
\label{DS_2}
The encoder-based methods, including PCNN (Piecewise CNN), max-pooling, multi-instance learning (MIL) modules, provide an infrastructure for distant supervision methods and reference for improvement of subsequent de-noise methods.

\textbf{PCNN+MIL}: Inspired by Zeng et al. \cite{zeng2014relation}, Zeng et al. \cite{zeng2015distant} exploit the PCNN with MIL, which divides the sentence into 3 segments based on the positions of two entities, to extract the relevant features automatically from a sentence and to get the important structural information.
And then, they deploy MIL to train the model with the highest confidence level instance to reduce the noise (used one sentence from one bag). This model outperforms several competitive baselines but ignores other instances in the bag.



\textbf{MIMLCNN}: The same purpose as above, Jiang et al. \cite{jiang2016relation} propose a multi-instance multi-label CNN for distant supervision and introduce assumption 3.
This work leverages CNN to extract features from a single sentence and then aggregates all sentence representations into an entity-pair-level representation by cross-sentence max-pooling. In this way, the model merges features from different sentences, not one sentence from one bag. Meanwhile, the multi-relation pattern between entity pairs can be considered. Hence, for a given entity pair, the model can predict multiple relations simultaneously.

\subsection{Representation-based Methods}



Unlike the previous simple encoder methods, Lin et al. \cite{lin2016neural,lin2017neural} introduce two novel attention models. The first uses the weight of the attention mechanism to flag all instances in one bag, and then apply the vector weighted sum of all the sentences to represent a bag. Hence, this model can identify important instances from noisy sentences, as well as utilizes all the information in the bag to optimize the performance. 
As a special case of the MIL, this model effectively reduces the influence of wrong labeled instances. The seconde one considers and leverages multi-lingual corpus, which is based on \cite{lin2016neural}. This work proposes mono-lingual attention and a cross-lingual attention mechanism to excavate diverse information hidden in the data of different languages, and the experimental results show that this work effectively model relation patterns among different languages. However, if each language builds a cross attention matrix, it doesn't seem realistic.

Yang et al. \cite{yang2017multi} and Du et al.'s \cite{du2018multi} work are similar, both of them use two different kinds of attention mechanism (two-level) to extract features. The first one is used to present the sentence, which looks like \cite{zhou2016attention}, and alleviates the distractions of irrelevant words. The second one is like \cite{lin2016neural}, leveraging attention mechanism to choose the weight of sentence with the highest probability.

In addition to using attention mechanisms, there are also ideas for incorporating more information to enhance the model. Banerjee et al. \cite{banerjee2018relation} propose a simple co-occurrence based strategy method for calculating the highest confidence in distant supervision bag, which uses the most frequent samples in the bag as the training label of the package. This is also a way to enhance the presentation.



\subsection{Knowledge-based Methods}
Only with the entity pair and context, the extracted features are not enough to represent the sentence, which may result in the performance in a low accuracy. In recent years, with the development of word vectors and knowledge graphs, additional background information has been introduced in RE, improving the external description of relation vectors \cite{ji2017distant,wang2018label,vashishth2018reside}.

In Ji et al.'s \cite{ji2017distant} work, they continue to choice the classic PCNN model to get sentence feature vectors. But in the data source processing step, this work introduces the conception of the relation vector from the knowledge graph $r = e_1 - e_2$ to present the relation, and then, combines these two kinds of vectors to a new vector by concatenation.
With the new vector, this model generates an attention weight vector to weight sum all of the sentence feature vectors as the bag's features. Meanwhile, to get additional background information, a description for entities from Freebase and Wikipedia pages is introduced to improve the entity representations. The experimental results show that this method contains more background knowledge to entities.

Following the above work \cite{ji2017distant}, Wang et al. \cite{wang2018label} also introduce the concept of the relation vector of knowledge graph $r = e_1 - e_2$ to present the relation. But this model is different from the above, it leverages the labeled data from the sentence itself, which means that all the labels of the training data are determined by sentence and aligned entity pairs. In the end, the sentences, forming a sentence pattern, can be classified into a diverse group by the types of aligned entities in the knowledge graph. Hence, this method also alleviates the wrong label problem and makes full use of the training corpus produced by distant supervision.

Since not all entities have description information, the above two methods may not apply to all entities, but some side information describing entities type or entity relations can be utilized. In Vashishth et al.'s \cite{vashishth2018reside} work, they consider some relevant and additional side information of KB, such as entity type and relation alias, and employ GCN to encode syntactic information from text. The entity type information, in this method, is integrated into an embedding to represent various entity types. And the relation alias come from some NLP toolkits (Stanford Open IE \cite{angeli2015leveraging}) or Paraphrase database PPDB \cite{pavlick2015ppdb}. This kind of description can also be seen as a knowledge improving the performance of RE model.

\subsection{Plug-and-play based Methods}
Either the "enhanced representation" or "external knowledge" methods mentioned above are to improve RE model itself, which is not universal. Hence, to directly construct a method as a plug-and-play component in distant supervision to reduce the noise of data sets must be a new insight.

\begin{figure*}[!t]
\centering \makeatletter\IfFileExists{qin2018dsgan.pdf}{\includegraphics[width=1.0\textwidth]{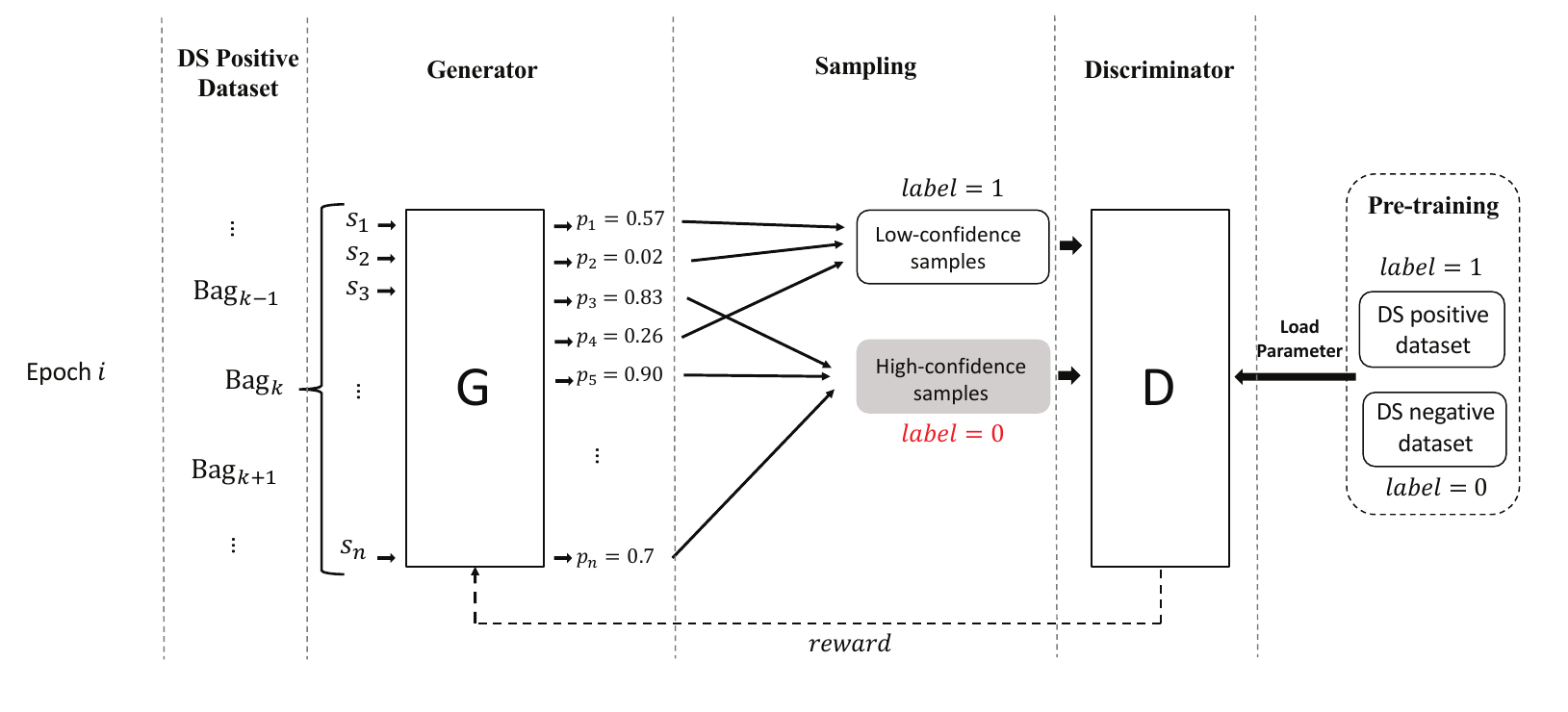}}{}
\makeatother
\caption{The structure of model DSGAN \cite{qin2018dsgan}.}
\label{lab_dsgan}
\end{figure*}

Qin et al. \cite{qin2018robust,qin2018dsgan} offer two approaches to constructing this plug-and-play component: One is the reinforcement learning framework for distant supervision \cite{qin2018robust} to solve false-positive case problem. The author argues that those wrong label sentences must be filtered by a hard decision, not by a soft weight of attention. In this way, this model generates less noise training data sets used in any previous state-of-the-art model. The second is DSGAN, using the idea of GAN to obtain a generator to classify the positive and negative samples in a bag from distant supervision. This is a kind of adversarial learning strategy, which could detect true-positive samples from the noisy distant supervision data sets. To some extent, these two methods alleviate the wrong label problem and form a new high-confidence training data sets. With this new data sets, some recent state-of-the-art models achieve further improvement in the experiment. Hence, both of the "reinforcement learning framework" and DSGAN can be seen as a plug-and-play component in distant supervision. The structure of DSGAN is shown in Figure \ref{lab_dsgan}.

\subsection{Discussion}

To summarize, distant supervision adopts various methods to get an excellent result. Figure \ref{envolution_of_DS} and Table \ref{tab:table_DS} illustrate that most of these works focus on de-noise algorithm. These works document that although distant supervision brings substantial benefits, its drawbacks, the noise in the data set, are also obvious. The above three methods (representation-based, knowledge-based, and plug-and-play based) cope the noise problem in different aspects, while they are still not optimal. Especially for knowledge-based methods,
introducing external knowledge will enable models with the ability of commonsense reasoning, which are missing in RE task. Meanwhile, fewer models devote to feature extraction, which means that the neural network variants have failed to significantly improve performance in pure text feature extraction. Consequently, no matter to improve the sentence encoder or de-noise algorithm, they only increase something extra burden to the basic model instead of solving insufficient training corpus problem (We compare the characteristics of supervision and distant supervision methods in Table \ref{comparisonMethods}). And of course, some researchers are considering other solutions, and we'll talk about those methods later.

\begin{table*}[]
\resizebox{\textwidth}{!}{
\begin{threeparttable}
\centering
\caption{The comparison of supervised methods and distant supervision methods.}
\begin{tabular}{cccc}
\toprule
   \textbf{Item}                        &\textbf{Features}          &\textbf{Supervision}            & \textbf{Distant Supervision}            \\
\midrule
    \multirow{4}[0]{*}{Data sets}       & Annotate mode             & Manually annotated             & Distant alignment KB        \\
                                        & Accuracy of labeled data  & High accuracy       & Low accuracy                                       \\
                                        & Noise                     & Low noise           & High noise                                         \\
                                        & Data Size                 & Small                          & Large                                   \\
\midrule
    \multirow{2}[0]{*}{Applicability}   &Model portability          & Low                            & High                                    \\
                                        &Cross-domain applicability & Low                           & High                                      \\
\midrule
    Accuracy of prediction              &        -          & High                           & Low                                              \\
\bottomrule
\end{tabular}
\end{threeparttable}
}
\label{comparisonMethods}
\end{table*}

\section{Other DNN-based RE}
\label{Other DNN-based RE}
All the above RE methods are to extract one relation of one entity pair. But in fact, one entity pair may have multiple relations in a sentence. To address this issue, Zhang et al. \cite{zhang2018attention} come up with a RE approach based on capsule networks with an attention mechanism, which outputs multi-relations of one entity pair. To some extent, this method considers more detailed features.
In addition, joint extraction methods \cite{ye2016jointly,miwa2016end,li2017neural,zheng2017joint,xiao2020joint}, known as end-to-end extraction, is also a new sight. Unlike the methods discussed earlier, the joint extraction method integrates NER and RE into one task, which outputs the relation and entity pair together.
In general, although the joint extraction method reduces the possibility of error propagation. compared with other methods, its accuracy and usability still have a large room for further improvement.

\section{Data Sets}
\label{DataSet}

\begin{table}[]
\centering
\caption{The comparison of data sets.}
\begin{tabular}{cccc}
\toprule
   \textbf{Data set}                     &\textbf{Relations}         &\begin{tabular}[c]{@{}l@{}}\textbf{Data Amount}\\ (words/language)\end{tabular}              \\
\midrule
    SemEval-2010 Task8                  & 18                        & 10717                           \\
    ACE04                               & 24                        & 350K                            \\
    NYT+Freebase                        & 53                        & 695059                       \\
    FewRel                              & 100                       & 70000                     \\
\bottomrule
\end{tabular}
\label{comparisonDatasets}
\end{table}

Variety of data sets exit for different methods, common data sets are SemEval 2010-task8 \cite{hendrickx2009semeval}, ACE series 2003-2005, NYT+Freebase \cite{riedel2010modeling}. The SemEval 2010-task8 and ACE series are commonly used for supervised learning classification tasks, while the NYT+Freebase for distant supervision methods. Table \ref{comparisonDatasets} shows the comparison of 4 representative data sets. The following is a brief introduction to these data sets.

\textbf{SemEval 2010-task8} is released in 2010, as an improvement on the Semeval 2007-task4, it provides a standard testbed for evaluating various methods of RE. This data set is used widely for evaluation, which contains 9 directional relations and an additional 'other' relation, resulting in 19 relation classes in total. Most of the supervised methods discussed in this article use the same data set.
The relations are as follows:

\begin{table*}[]
\centering
\caption{Some examples from SemEval 2010-task8.}
    \begin{tabular*}{\textwidth}{lp{0.8\textwidth}}
        \toprule
        Example1: & \textless{}e1\textgreater{}People\textless{}/e1\textgreater have been moving back into \textless{}e2\textgreater{}downtown\textless{}/e2\textgreater{}.                                                                         \\
        Relation: & Entity-Destination(e1,e2).                                                                                                                                                                                                      \\ \midrule
        Example2: & Cieply's \textless{}e1\textgreater{}story\textless{}/e1\textgreater makes a compelling \textless{}e2\textgreater{}point\textless{}/e2\textgreater about modern-day studio economics. \\
        Relation: & Message-Topic(e1,e2).                                                                                                                                                                                                           \\ \bottomrule
        \label{tab3}
    \end{tabular*}
\end{table*}

\begin{itemize}
\item Cause-Effect: An event or object leads to an effect. (those cancers were caused by radiation exposures)
\item Component-Whole: An object is a component of a larger whole. (my apartment has a large kitchen)
\item Content-Container: An object is physically stored in a delineated area of space, the container. (Earth is located in the Milky Way)
\item Entity-Destination: An entity is moving towards a destination. (the boy went to bed)
\item Entity-Origin: An entity is coming or is derived from an origin, e.g., position or material. (letters from foreign countries)
\item Message-Topic: An act of communication, written or spoken, is about a topic. (the lecture was about semantics)
\item Member-Collection: A member forms a nonfunctional part of a collection. (there are many trees in the forest)
\item Instrument-Agency: An agent uses an instrument. (phone operator)
\item Product-Agency: A producer causes a product to exist. (a factory manufactures suits)
\item Other: If none of the above nine relations appears to be suitable.
\end{itemize}
This data set contains 10717 labeled data, of which 8000 are used for training and the remaining 2717 for testing. The nominals of the sentences in the data set are marked by XML tags. Table \ref{tab3} shows an example of labeled sentences.

\begin{table}[]
\centering
\caption{An introduction to ACE data sets.}
\begin{tabular}{llll}
\toprule
    \textbf{Corpus}   & \begin{tabular}[c]{@{}l@{}}\textbf{Data Amount}\\ (words/language)\end{tabular} & \textbf{Tasks} & \textbf{Languages} \\
\midrule
    ACE03 & \begin{tabular}[c]{@{}l@{}}100K training\\ 50K evaluation\end{tabular} & \begin{tabular}[c]{@{}l@{}}entities\\ relations\end{tabular} & \begin{tabular}[c]{@{}l@{}}Chinese\\ English\\ Arabic\end{tabular} \\
    ACE04 & \begin{tabular}[c]{@{}l@{}}300K training\\ 50K evaluation\end{tabular} & \begin{tabular}[c]{@{}l@{}}entities\\ relations \\events \end{tabular} & \begin{tabular}[c]{@{}l@{}}Chinese\\ English\\ Arabic\end{tabular} \\
    ACE05 & \begin{tabular}[c]{@{}l@{}}750K training\\ 150K evaluation\end{tabular} & \begin{tabular}[c]{@{}l@{}}entities\\ relations \\events \end{tabular} & \begin{tabular}[c]{@{}l@{}}Chinese\\ English\\ Arabic\end{tabular} \\
\bottomrule
\end{tabular}
\label{tab4}
\end{table}

\textbf{ACE2003-2005 Series} come from LDC (Linguistic Data consortium), consisting of various types of annotated for entities and relations. In three different years of the corpus (ACE 2003, 2004, 2005), the ACE tasks are more complex than SemEval 2010-task8. The corpus can be divided into broadcast news, newswire and telephone conversations, including the complete set of English, Arabic, Chinese training data. So the annotated entities in this corpus are pronoun words or other irregular words. In addition to the task of RE, they can also be applied in the following five tasks: Entity Detection and Recognition, Entity Mention Detection, EDR Co-reference, Relation Mention Detection, and Relation Detection and Recognition of given reference entities. This data set is typically used in supervised models. Table \ref{tab4} shows the introduction of ACE data sets.

\textbf{NYT+Freebase} \label{section:NYT+Freebase} means New York times + Freebase, which is a common way to generate data sets in distant supervision. The distant supervision method produces data sets by extracting relations through the heuristic alignment of text and entities in the KB (talked about that in section \ref{Traditional method}). As a result, in this way, distant supervision creates large-scale training data automatically. This process is shown in Figure \ref{fig_ds}. The most widely used data set is generated by Riedel et al. \cite{riedel2010modeling}, which has 522611 training sentences and 172448 test sentences, labeled by 53 candidate relations in Freebase, and an extra-label of NA (nearly 80\% of the sentences in the training data are labeled as NA).

\textbf{Other Data sets} \label{other datasets} include SemEval2017-task10 \cite{bethard2016proceedings}, SemEval2018-task7 \cite{gabor2018semeval}, TACRED \cite{zhang2017tacred}, KBP37 \cite{zhang2015relation}, DDIExtraction2011 \cite{segura20111st}, DDIExtraction2013 \cite{segura2013semeval},
 FewRel \cite{han2018fewrel}, FewRel2 \cite{gao2019fewrel}.


\section{Conclusions}
\label{conclusions}
This article begins by laying out the general framework and some basic concepts, focusing on DNN-based methods in supervised and distant supervision.
Based on the above elaboration, both supervised and distant supervision methods have their own characteristics. Supervised methods are better suited for specific domain, while distant supervision methods are better for generic domains. As a result, it is difficult to specify which methods are currently the best. Hence, we just compare the characteristics of supervised and distant supervised methods in Table \ref{comparisonMethods}. In general, despite its long success, RE methods still have several problems as follow:

\textbf{Transfer learning}: is a cross-domain adaptive solution. It provides better domain transfer capability for RE model, especially in supervised learning method, making the model have better extensibility to another corpus. Presently, world knowledge and the number of relations are not stagnant, creating a dynamic knowledge base. Nevertheless, most of RE models, with low utilization of dynamic knowledge, are explored in predefined relation inventory and suffered from the insufficient data sets. To keep the model continuously receiving new training samples and making full use of other related corpus is worthy of exploration. It is noted that the results in this field are not yet significant \cite{di2019relation,sun2019distantly,zhang2019transfer}.

\textbf{Relationship Reasoning}: presents a way to synthesize background knowledge and be aware of new knowledge based on logical ability. For RE, with the reasoning ability and some existing relational facts, we may use this logical ability to expand our existing knowledge. At the same time, in the open domain, without the relation inventory in advance, it is a very forward-looking work to obtain the relation directly from the existing corpus in the real world. Hence, using background knowledge to obtain the relation by reasoning is an exciting way to solve the problem. Several attempts \cite{ji2017distant,wang2018label,ren2018neural} have been made to use a knowledge graph or introduce entity description information to RE. However, how to further improve the reasoning mechanism still needs studying.

\textbf{Relationship Framework}: limits RE task in an established framework. Another problem is the diversity of relation inventory in data sets (out-of-vocabulary problem). Different data sets have different definitions of relation inventories, making the training model of each data set limited in domain adaptability. If the industry can construct a framework with an unified description of all kinds of relations and make RE have a hierarchical structure, then it may have a better multi-domain adaptation. Predecessors have done a lot of work on the definition of relation inventories, but no agreement has been reached \cite{hendrickx2009semeval}.

\textbf{Cross-sentence RE}: Moreover, cross-sentence RE is likewise one significant field. The current RE task mainly focuses on processing the entity pair relation in intra-sentence. In the actual situation, most entities in the text represent the relation between each other by multi sentences. Current models may not be able to handle such tasks directly. Especially, distant supervision generates a large scale of document-level corpora, which can only be solved by a cross-sentence RE technique.
some researchers \cite{sahu2019inter,Guo2019,Zhang2019} propose a series of novel methods to obtain the entity relation across sentences. These studies offer some important insights into cross-sentence RE.

\textbf{Others}: In addition to these fields of research, several problems in the existing methods also need solving, one is the problem of error propagation in supervised methods, and the other is the problem of wrong label in distant supervision. Especially for the latter one, if the wrong label problem can be well solved, a large amount of effective sample data will be obtained, which will make an important contribution to this field.

\begin{acknowledgements}
This work was supported by National Natural Science Foundation of China (No. U19A2059), and by Ministry of Science and Technology of Sichuan Province Program (NO.2021YFG0018\&No.20ZDYF0343).

We sincerely thank Mr. Kombou Victor, Anto Leoba Jonathan, Rufai Yusuf Zakri and Owusu Wilson Jim for their helpful discussions.

\end{acknowledgements}


%
%

\bibliographystyle{spbasic-unsort}      
\bibliography{sample}   

\begin{thebibliography}{87}
\providecommand{\natexlab}[1]{#1}
\providecommand{\url}[1]{{#1}}
\providecommand{\urlprefix}{URL }
\expandafter\ifx\csname urlstyle\endcsname\relax
  \providecommand{\doi}[1]{DOI~\discretionary{}{}{}#1}\else
  \providecommand{\doi}{DOI~\discretionary{}{}{}\begingroup
  \urlstyle{rm}\Url}\fi
\providecommand{\eprint}[2][]{\url{#2}}

\bibitem[{Kumar(2017)}]{kumar2017survey}
Kumar S (2017) {A survey of deep learning methods for relation extraction}.
  arXiv preprint arXiv:170503645

\bibitem[{Pawar et~al.(2017)Pawar, Palshikar, and
  Bhattacharyya}]{pawar2017relation}
Pawar S, Palshikar GK, Bhattacharyya P (2017) Relation extraction: A survey.
  arXiv preprint arXiv:171205191

\bibitem[{Hendrickx et~al.(2009)Hendrickx, Kim, Kozareva, Nakov,
  {\'O}~S{\'e}aghdha, Pad{\'o}, Pennacchiotti, Romano, and
  Szpakowicz}]{hendrickx2009semeval}
Hendrickx I, Kim SN, Kozareva Z, Nakov P, {\'O}~S{\'e}aghdha D, Pad{\'o} S,
  Pennacchiotti M, Romano L, Szpakowicz S (2009) Semeval-2010 task 8: Multi-way
  classification of semantic relations between pairs of nominals. In:
  Proceedings of the Workshop on Semantic Evaluations: Recent Achievements and
  Future Directions, Association for Computational Linguistics, pp 94--99

\bibitem[{Han et~al.(2018)Han, Zhu, Yu, Wang, Yao, Liu, and
  Sun}]{han2018fewrel}
Han X, Zhu H, Yu P, Wang Z, Yao Y, Liu Z, Sun M (2018) Fewrel: A large-scale
  supervised few-shot relation classification dataset with state-of-the-art
  evaluation. arXiv preprint arXiv:181010147

\bibitem[{Gao et~al.(2019)Gao, Han, Zhu, Liu, Li, Sun, and
  Zhou}]{gao2019fewrel}
Gao T, Han X, Zhu H, Liu Z, Li P, Sun M, Zhou J (2019) Fewrel 2.0: Towards more
  challenging few-shot relation classification. arXiv preprint arXiv:191007124

\bibitem[{Riedel et~al.(2010)Riedel, Yao, and McCallum}]{riedel2010modeling}
Riedel S, Yao L, McCallum A (2010) Modeling relations and their mentions
  without labeled text. In: Joint European Conference on Machine Learning and
  Knowledge Discovery in Databases, Springer, pp 148--163

\bibitem[{LeCun et~al.(1989)LeCun, Boser, Denker, Henderson, Howard, Hubbard,
  and Jackel}]{lecun1989backpropagation}
LeCun Y, Boser B, Denker JS, Henderson D, Howard RE, Hubbard W, Jackel LD
  (1989) Backpropagation applied to handwritten zip code recognition. Neural
  computation 1(4):541--551

\bibitem[{Elman(1991)}]{elman1991distributed}
Elman JL (1991) Distributed representations, simple recurrent networks, and
  grammatical structure. Machine learning 7(2-3):195--225

\bibitem[{Socher et~al.(2012)Socher, Huval, Manning, and
  Ng}]{socher2012semantic}
Socher R, Huval B, Manning CD, Ng AY (2012) Semantic compositionality through
  recursive matrix-vector spaces. In: Proceedings of the 2012 joint conference
  on empirical methods in natural language processing and computational natural
  language learning, Association for Computational Linguistics, pp 1201--1211

\bibitem[{Scarselli et~al.(2008)Scarselli, Gori, Tsoi, Hagenbuchner, and
  Monfardini}]{scarselli2008graph}
Scarselli F, Gori M, Tsoi AC, Hagenbuchner M, Monfardini G (2008) The graph
  neural network model. IEEE Transactions on Neural Networks 20(1):61--80

\bibitem[{Zhang et~al.(2015)Zhang, Zheng, Hu, and
  Yang}]{zhang2015bidirectional}
Zhang S, Zheng D, Hu X, Yang M (2015) Bidirectional long short-term memory
  networks for relation classification. In: Proceedings of the 29th Pacific
  Asia conference on language, information and computation, pp 73--78

\bibitem[{Sundermeyer et~al.(2012)Sundermeyer, Schl{\"u}ter, and
  Ney}]{sundermeyer2012lstm}
Sundermeyer M, Schl{\"u}ter R, Ney H (2012) Lstm neural networks for language
  modeling. In: Thirteenth annual conference of the international speech
  communication association

\bibitem[{Hochreiter and Schmidhuber(1997)}]{hochreiter1997long}
Hochreiter S, Schmidhuber J (1997) Long short-term memory. Neural computation
  9(8):1735--1780

\bibitem[{Chung et~al.(2014)Chung, Gulcehre, Cho, and
  Bengio}]{chung2014empirical}
Chung J, Gulcehre C, Cho K, Bengio Y (2014) Empirical evaluation of gated
  recurrent neural networks on sequence modeling. arXiv preprint arXiv:14123555

\bibitem[{Mikolov et~al.(2013)Mikolov, Chen, Corrado, and
  Dean}]{mikolov2013efficient}
Mikolov T, Chen K, Corrado G, Dean J (2013) Efficient estimation of word
  representations in vector space. arXiv preprint arXiv:13013781

\bibitem[{Turian et~al.(2010)Turian, Ratinov, and Bengio}]{turian2010word}
Turian J, Ratinov L, Bengio Y (2010) Word representations: a simple and general
  method for semi-supervised learning. In: Proceedings of the 48th annual
  meeting of the association for computational linguistics, Association for
  Computational Linguistics, pp 384--394

\bibitem[{Pennington et~al.(2014)Pennington, Socher, and
  Manning}]{pennington2014glove}
Pennington J, Socher R, Manning C (2014) Glove: Global vectors for word
  representation. In: Proceedings of the 2014 conference on empirical methods
  in natural language processing (EMNLP), pp 1532--1543

\bibitem[{Zeng et~al.(2014)Zeng, Liu, Lai, Zhou, Zhao
  et~al.}]{zeng2014relation}
Zeng D, Liu K, Lai S, Zhou G, Zhao J, et~al. (2014) Relation classification via
  convolutional deep neural network

\bibitem[{Nguyen and Grishman(2015)}]{nguyen2015relation}
Nguyen TH, Grishman R (2015) Relation extraction: Perspective from
  convolutional neural networks. In: Proceedings of the 1st Workshop on Vector
  Space Modeling for Natural Language Processing, pp 39--48

\bibitem[{Santos et~al.(2015)Santos, Xiang, and Zhou}]{santos2015classifying}
Santos CNd, Xiang B, Zhou B (2015) Classifying relations by ranking with
  convolutional neural networks. arXiv preprint arXiv:150406580

\bibitem[{Wang et~al.(2016)Wang, Cao, De~Melo, and Liu}]{wang2016relation}
Wang L, Cao Z, De~Melo G, Liu Z (2016) Relation classification via multi-level
  attention cnns. In: Proceedings of the 54th Annual Meeting of the Association
  for Computational Linguistics (Volume 1: Long Papers), pp 1298--1307

\bibitem[{Zhang and Wang(2015)}]{zhang2015relation}
Zhang D, Wang D (2015) Relation classification via recurrent neural network.
  arXiv preprint arXiv:150801006

\bibitem[{Qin et~al.(2017)Qin, Xu, and Guo}]{qin2017designing}
Qin P, Xu W, Guo J (2017) Designing an adaptive attention mechanism for
  relation classification. In: 2017 International Joint Conference on Neural
  Networks (IJCNN), IEEE, pp 4356--4362

\bibitem[{Zhang et~al.(2019)Zhang, Cui, Gao, Nie, Xu, Yang, Xi, and
  Yin}]{zhang2019multi}
Zhang C, Cui C, Gao S, Nie X, Xu W, Yang L, Xi X, Yin Y (2019) Multi-gram
  cnn-based self-attention model for relation classification. IEEE Access
  7:5343--5357

\bibitem[{Ren et~al.(2018)Ren, Zhou, Liu, Li, Zhao, Liu, and
  Liang}]{ren2018neural}
Ren F, Zhou D, Liu Z, Li Y, Zhao R, Liu Y, Liang X (2018) Neural relation
  classification with text descriptions. In: Proceedings of the 27th
  International Conference on Computational Linguistics, pp 1167--1177

\bibitem[{Zhang and Xiang(2018)}]{zhang2018relation}
Zhang L, Xiang F (2018) Relation classification via bilstm-cnn. In:
  International Conference on Data Mining and Big Data, Springer, pp 373--382

\bibitem[{Mooney and Bunescu(2006)}]{mooney2006subsequence}
Mooney RJ, Bunescu RC (2006) Subsequence kernels for relation extraction. In:
  Advances in neural information processing systems, pp 171--178

\bibitem[{Xu et~al.(2015)Xu, Mou, Li, Chen, Peng, and Jin}]{xu2015classifying}
Xu Y, Mou L, Li G, Chen Y, Peng H, Jin Z (2015) Classifying relations via long
  short term memory networks along shortest dependency paths. In: proceedings
  of the 2015 conference on empirical methods in natural language processing,
  pp 1785--1794

\bibitem[{Cai et~al.(2016)Cai, Zhang, and Wang}]{cai2016bidirectional}
Cai R, Zhang X, Wang H (2016) Bidirectional recurrent convolutional neural
  network for relation classification. In: Proceedings of the 54th Annual
  Meeting of the Association for Computational Linguistics (Volume 1: Long
  Papers), vol~1, pp 756--765

\bibitem[{Guo et~al.(2019)Guo, Zhang, Yang, Xu, and Ye}]{guo2019single}
Guo X, Zhang H, Yang H, Xu L, Ye Z (2019) A single attention-based combination
  of cnn and rnn for relation classification. IEEE Access 7:12467--12475

\bibitem[{Jin et~al.(2020)Jin, Song, Zhang, Xu, Ma, and Yu}]{jin2020relation}
Jin L, Song L, Zhang Y, Xu K, Ma Wy, Yu D (2020) Relation extraction exploiting
  full dependency forests. In: Proceedings of the AAAI Conference on Artificial
  Intelligence, vol~34, pp 8034--8041

\bibitem[{Hearst(1992)}]{hearst1992automatic}
Hearst MA (1992) Automatic acquisition of hyponyms from large text corpora. In:
  Proceedings of the 14th conference on Computational linguistics-Volume 2,
  Association for Computational Linguistics, pp 539--545

\bibitem[{Berland and Charniak(1999)}]{berland1999finding}
Berland M, Charniak E (1999) Finding parts in very large corpora. In:
  Proceedings of the 37th annual meeting of the Association for Computational
  Linguistics

\bibitem[{Brin(1998)}]{brin1998extracting}
Brin S (1998) Extracting patterns and relations from the world wide web. In:
  International workshop on the world wide web and databases, Springer, pp
  172--183

\bibitem[{Agichtein and Gravano(2000)}]{agichtein2000snowball}
Agichtein E, Gravano L (2000) Snowball: Extracting relations from large
  plain-text collections. In: Proceedings of the fifth ACM conference on
  Digital libraries, ACM, pp 85--94

\bibitem[{Etzioni et~al.(2004)Etzioni, Cafarella, Downey, Kok, Popescu, Shaked,
  Soderland, Weld, and Yates}]{etzioni2004web}
Etzioni O, Cafarella M, Downey D, Kok S, Popescu AM, Shaked T, Soderland S,
  Weld DS, Yates A (2004) Web-scale information extraction in
  knowitall:(preliminary results). In: Proceedings of the 13th international
  conference on World Wide Web, ACM, pp 100--110

\bibitem[{Yates et~al.(2007)Yates, Cafarella, Banko, Etzioni, Broadhead, and
  Soderland}]{yates2007textrunner}
Yates A, Cafarella M, Banko M, Etzioni O, Broadhead M, Soderland S (2007)
  Textrunner: open information extraction on the web. In: Proceedings of Human
  Language Technologies: The Annual Conference of the North American Chapter of
  the Association for Computational Linguistics: Demonstrations, Association
  for Computational Linguistics, pp 25--26

\bibitem[{Phi et~al.(2018)Phi, Santoso, Shimbo, and Matsumoto}]{phi2018ranking}
Phi VT, Santoso J, Shimbo M, Matsumoto Y (2018) Ranking-based automatic seed
  selection and noise reduction for weakly supervised relation extraction. In:
  Proceedings of the 56th Annual Meeting of the Association for Computational
  Linguistics (Volume 2: Short Papers), pp 89--95

\bibitem[{Hasegawa et~al.(2004)Hasegawa, Sekine, and
  Grishman}]{hasegawa2004discovering}
Hasegawa T, Sekine S, Grishman R (2004) Discovering relations among named
  entities from large corpora. In: Proceedings of the 42nd annual meeting on
  association for computational linguistics, Association for Computational
  Linguistics, p 415

\bibitem[{Rink and Harabagiu(2010)}]{rink2010utd}
Rink B, Harabagiu S (2010) Utd: Classifying semantic relations by combining
  lexical and semantic resources. In: Proceedings of the 5th International
  Workshop on Semantic Evaluation, Association for Computational Linguistics,
  pp 256--259

\bibitem[{Kambhatla(2004)}]{kambhatla2004combining}
Kambhatla N (2004) Combining lexical, syntactic, and semantic features with
  maximum entropy models for extracting relations. In: Proceedings of the ACL
  2004 on Interactive poster and demonstration sessions, Association for
  Computational Linguistics, p~22

\bibitem[{Bunescu and Mooney(2005)}]{bunescu2005shortest}
Bunescu RC, Mooney RJ (2005) A shortest path dependency kernel for relation
  extraction. In: Proceedings of the conference on human language technology
  and empirical methods in natural language processing, Association for
  Computational Linguistics, pp 724--731

\bibitem[{Mintz et~al.(2009)Mintz, Bills, Snow, and
  Jurafsky}]{mintz2009distant}
Mintz M, Bills S, Snow R, Jurafsky D (2009) Distant supervision for relation
  extraction without labeled data. In: Proceedings of the Joint Conference of
  the 47th Annual Meeting of the ACL and the 4th International Joint Conference
  on Natural Language Processing of the AFNLP: Volume 2-Volume 2, Association
  for Computational Linguistics, pp 1003--1011

\bibitem[{Manning et~al.(2014)Manning, Surdeanu, Bauer, Finkel, Bethard, and
  McClosky}]{manning2014stanford}
Manning C, Surdeanu M, Bauer J, Finkel J, Bethard S, McClosky D (2014) The
  stanford corenlp natural language processing toolkit. In: Proceedings of 52nd
  annual meeting of the association for computational linguistics: system
  demonstrations, pp 55--60

\bibitem[{Liu et~al.(2013)Liu, Sun, Chao, and Che}]{liu2013convolution}
Liu C, Sun W, Chao W, Che W (2013) Convolution neural network for relation
  extraction. In: International Conference on Advanced Data Mining and
  Applications, Springer, pp 231--242

\bibitem[{Kim(2014)}]{kim2014convolutional}
Kim Y (2014) Convolutional neural networks for sentence classification. arXiv
  preprint arXiv:14085882

\bibitem[{Kalchbrenner et~al.(2014)Kalchbrenner, Grefenstette, and
  Blunsom}]{kalchbrenner2014convolutional}
Kalchbrenner N, Grefenstette E, Blunsom P (2014) A convolutional neural network
  for modelling sentences. arXiv preprint arXiv:14042188

\bibitem[{Xu et~al.(2015)Xu, Feng, Huang, and Zhao}]{xu2015semantic}
Xu K, Feng Y, Huang S, Zhao D (2015) Semantic relation classification via
  convolutional neural networks with simple negative sampling. arXiv preprint
  arXiv:150607650

\bibitem[{Zhou et~al.(2016)Zhou, Shi, Tian, Qi, Li, Hao, and
  Xu}]{zhou2016attention}
Zhou P, Shi W, Tian J, Qi Z, Li B, Hao H, Xu B (2016) Attention-based
  bidirectional long short-term memory networks for relation classification.
  In: Proceedings of the 54th Annual Meeting of the Association for
  Computational Linguistics (Volume 2: Short Papers), vol~2, pp 207--212

\bibitem[{Xu et~al.(2016)Xu, Jia, Mou, Li, Chen, Lu, and Jin}]{xu2016improved}
Xu Y, Jia R, Mou L, Li G, Chen Y, Lu Y, Jin Z (2016) Improved relation
  classification by deep recurrent neural networks with data augmentation.
  arXiv preprint arXiv:160103651

\bibitem[{Xiao and Liu(2016)}]{xiao2016semantic}
Xiao M, Liu C (2016) Semantic relation classification via hierarchical
  recurrent neural network with attention. In: Proceedings of COLING 2016, the
  26th International Conference on Computational Linguistics: Technical Papers,
  pp 1254--1263

\bibitem[{Lee et~al.(2019)Lee, Seo, and Choi}]{lee2019semantic}
Lee J, Seo S, Choi YS (2019) Semantic relation classification via bidirectional
  lstm networks with entity-aware attention using latent entity typing. arXiv
  preprint arXiv:190108163

\bibitem[{Zheng et~al.(2016)Zheng, Xu, Zhou, Bao, Qi, and Xu}]{zheng2016neural}
Zheng S, Xu J, Zhou P, Bao H, Qi Z, Xu B (2016) A neural network framework for
  relation extraction: Learning entity semantic and relation pattern.
  Knowledge-Based Systems 114:12--23

\bibitem[{Wang et~al.(2020)Wang, Qin, Lu, Luo, and Liu}]{wang2020direction}
Wang H, Qin K, Lu G, Luo G, Liu G (2020) Direction-sensitive relation
  extraction using bi-sdp attention model. Knowledge-Based Systems p 105928

\bibitem[{Zhang et~al.(2020)Zhang, Shu, Yu, Liu, Zhao, Li, and
  Guo}]{zhang2020distilling}
Zhang Z, Shu X, Yu B, Liu T, Zhao J, Li Q, Guo L (2020) Distilling knowledge
  from well-informed soft labels for neural relation extraction. In: AAAI, pp
  9620--9627

\bibitem[{Lyu et~al.(2020)Lyu, Cheng, Wu, Cui, Chen, and
  Miao}]{lyu2020auxiliary}
Lyu S, Cheng J, Wu X, Cui L, Chen H, Miao C (2020) Auxiliary learning for
  relation extraction. IEEE Transactions on Emerging Topics in Computational
  Intelligence

\bibitem[{Zeng et~al.(2015)Zeng, Liu, Chen, and Zhao}]{zeng2015distant}
Zeng D, Liu K, Chen Y, Zhao J (2015) Distant supervision for relation
  extraction via piecewise convolutional neural networks. In: Proceedings of
  the 2015 Conference on Empirical Methods in Natural Language Processing, pp
  1753--1762

\bibitem[{Jiang et~al.(2016)Jiang, Wang, Li, and Wang}]{jiang2016relation}
Jiang X, Wang Q, Li P, Wang B (2016) Relation extraction with multi-instance
  multi-label convolutional neural networks. In: Proceedings of COLING 2016,
  the 26th International Conference on Computational Linguistics: Technical
  Papers, pp 1471--1480

\bibitem[{Yang et~al.(2017)Yang, Ng, Mooney, and Dong}]{yang2017multi}
Yang L, Ng TLJ, Mooney C, Dong R (2017) Multi-level attention-based neural
  networks for distant supervised relation extraction. In: AICS, pp 206--218

\bibitem[{Lin et~al.(2017)Lin, Liu, and Sun}]{lin2017neural}
Lin Y, Liu Z, Sun M (2017) Neural relation extraction with multi-lingual
  attention. In: Proceedings of the 55th Annual Meeting of the Association for
  Computational Linguistics (Volume 1: Long Papers), pp 34--43

\bibitem[{Lin et~al.(2016)Lin, Shen, Liu, Luan, and Sun}]{lin2016neural}
Lin Y, Shen S, Liu Z, Luan H, Sun M (2016) Neural relation extraction with
  selective attention over instances. In: Proceedings of the 54th Annual
  Meeting of the Association for Computational Linguistics (Volume 1: Long
  Papers), vol~1, pp 2124--2133

\bibitem[{Banerjee and Tsioutsiouliklis(2018)}]{banerjee2018relation}
Banerjee S, Tsioutsiouliklis K (2018) Relation extraction using multi-encoder
  lstm network on a distant supervised dataset. In: 2018 IEEE 12th
  International Conference on Semantic Computing (ICSC), IEEE, pp 235--238

\bibitem[{Du et~al.(2018)Du, Han, Way, and Wan}]{du2018multi}
Du J, Han J, Way A, Wan D (2018) Multi-level structured self-attentions for
  distantly supervised relation extraction. arXiv preprint arXiv:180900699

\bibitem[{Ji et~al.(2017)Ji, Liu, He, and Zhao}]{ji2017distant}
Ji G, Liu K, He S, Zhao J (2017) Distant supervision for relation extraction
  with sentence-level attention and entity descriptions. In: Thirty-First AAAI
  Conference on Artificial Intelligence

\bibitem[{Wang et~al.(2018)Wang, Zhang, Wang, Zhou, Chen, Zhang, Zhu, and
  Chen}]{wang2018label}
Wang G, Zhang W, Wang R, Zhou Y, Chen X, Zhang W, Zhu H, Chen H (2018)
  Label-free distant supervision for relation extraction via knowledge graph
  embedding. In: Proceedings of the 2018 Conference on Empirical Methods in
  Natural Language Processing, pp 2246--2255

\bibitem[{Vashishth et~al.(2018)Vashishth, Joshi, Prayaga, Bhattacharyya, and
  Talukdar}]{vashishth2018reside}
Vashishth S, Joshi R, Prayaga SS, Bhattacharyya C, Talukdar P (2018) Reside:
  Improving distantly-supervised neural relation extraction using side
  information. In: Proceedings of the 2018 Conference on Empirical Methods in
  Natural Language Processing, pp 1257--1266

\bibitem[{Qin et~al.(2018{\natexlab{a}})Qin, Xu, and Wang}]{qin2018robust}
Qin P, Xu W, Wang WY (2018{\natexlab{a}}) Robust distant supervision relation
  extraction via deep reinforcement learning. arXiv preprint arXiv:180509927

\bibitem[{Qin et~al.(2018{\natexlab{b}})Qin, Xu, and Wang}]{qin2018dsgan}
Qin P, Xu W, Wang WY (2018{\natexlab{b}}) Dsgan: Generative adversarial
  training for distant supervision relation extraction. arXiv preprint
  arXiv:180509929

\bibitem[{Angeli et~al.(2015)Angeli, Premkumar, and
  Manning}]{angeli2015leveraging}
Angeli G, Premkumar MJJ, Manning CD (2015) Leveraging linguistic structure for
  open domain information extraction. In: Proceedings of the 53rd Annual
  Meeting of the Association for Computational Linguistics and the 7th
  International Joint Conference on Natural Language Processing (Volume 1: Long
  Papers), vol~1, pp 344--354

\bibitem[{Pavlick et~al.(2015)Pavlick, Rastogi, Ganitkevitch, Van~Durme, and
  Callison-Burch}]{pavlick2015ppdb}
Pavlick E, Rastogi P, Ganitkevitch J, Van~Durme B, Callison-Burch C (2015) Ppdb
  2.0: Better paraphrase ranking, fine-grained entailment relations, word
  embeddings, and style classification. In: Proceedings of the 53rd Annual
  Meeting of the Association for Computational Linguistics and the 7th
  International Joint Conference on Natural Language Processing (Volume 2:
  Short Papers), vol~2, pp 425--430

\bibitem[{Zhang et~al.(2018)Zhang, Deng, Sun, Chen, Zhang, and
  Chen}]{zhang2018attention}
Zhang N, Deng S, Sun Z, Chen X, Zhang W, Chen H (2018) Attention-based capsule
  networks with dynamic routing for relation extraction. arXiv preprint
  arXiv:181211321

\bibitem[{Ye et~al.(2016)Ye, Chao, Luo, and Li}]{ye2016jointly}
Ye H, Chao W, Luo Z, Li Z (2016) Jointly extracting relations with class ties
  via effective deep ranking. arXiv preprint arXiv:161207602

\bibitem[{Miwa and Bansal(2016)}]{miwa2016end}
Miwa M, Bansal M (2016) End-to-end relation extraction using lstms on sequences
  and tree structures. arXiv preprint arXiv:160100770

\bibitem[{Li et~al.(2017)Li, Zhang, Fu, and Ji}]{li2017neural}
Li F, Zhang M, Fu G, Ji D (2017) A neural joint model for entity and relation
  extraction from biomedical text. BMC bioinformatics 18(1):198

\bibitem[{Zheng et~al.(2017)Zheng, Wang, Bao, Hao, Zhou, and
  Xu}]{zheng2017joint}
Zheng S, Wang F, Bao H, Hao Y, Zhou P, Xu B (2017) Joint extraction of entities
  and relations based on a novel tagging scheme. arXiv preprint arXiv:170605075

\bibitem[{Xiao et~al.(2020)Xiao, Tan, Fan, Xu, and Zhu}]{xiao2020joint}
Xiao Y, Tan C, Fan Z, Xu Q, Zhu W (2020) Joint entity and relation extraction
  with a hybrid transformer and reinforcement learning based model. In: AAAI,
  pp 9314--9321

\bibitem[{Bethard et~al.(2016)Bethard, Carpuat, Cer, Jurgens, Nakov, and
  Zesch}]{bethard2016proceedings}
Bethard S, Carpuat M, Cer D, Jurgens D, Nakov P, Zesch T (2016) Proceedings of
  the 10th international workshop on semantic evaluation (semeval-2016). In:
  Proceedings of the 10th International Workshop on Semantic Evaluation
  (SemEval-2016)

\bibitem[{G{\'a}bor et~al.(2018)G{\'a}bor, Buscaldi, Schumann, QasemiZadeh,
  Zargayouna, and Charnois}]{gabor2018semeval}
G{\'a}bor K, Buscaldi D, Schumann AK, QasemiZadeh B, Zargayouna H, Charnois T
  (2018) Semeval-2018 task 7: Semantic relation extraction and classification
  in scientific papers. In: Proceedings of The 12th International Workshop on
  Semantic Evaluation, pp 679--688

\bibitem[{Zhang et~al.(2017)Zhang, Zhong, Chen, Angeli, and
  Manning}]{zhang2017tacred}
Zhang Y, Zhong V, Chen D, Angeli G, Manning CD (2017) Position-aware attention
  and supervised data improve slot filling. In: Proceedings of the 2017
  Conference on Empirical Methods in Natural Language Processing (EMNLP 2017),
  pp 35--45, \urlprefix\url{https://nlp.stanford.edu/pubs/zhang2017tacred.pdf}

\bibitem[{Segura~Bedmar et~al.(2011)Segura~Bedmar, Martinez, and
  S{\'a}nchez~Cisneros}]{segura20111st}
Segura~Bedmar I, Martinez P, S{\'a}nchez~Cisneros D (2011) The 1st
  ddiextraction-2011 challenge task: Extraction of drug-drug interactions from
  biomedical texts

\bibitem[{Segura~Bedmar et~al.(2013)Segura~Bedmar, Mart{\'\i}nez, and
  Herrero~Zazo}]{segura2013semeval}
Segura~Bedmar I, Mart{\'\i}nez P, Herrero~Zazo M (2013) Semeval-2013 task 9:
  Extraction of drug-drug interactions from biomedical texts (ddiextraction
  2013). Association for Computational Linguistics

\bibitem[{Di et~al.(2019)Di, Shen, and Chen}]{di2019relation}
Di S, Shen Y, Chen L (2019) Relation extraction via domain-aware transfer
  learning. In: Proceedings of the 25th ACM SIGKDD International Conference on
  Knowledge Discovery \& Data Mining, pp 1348--1357

\bibitem[{Sun and Wu(2019)}]{sun2019distantly}
Sun C, Wu Y (2019) Distantly supervised entity relation extraction with adapted
  manual annotations. In: Proceedings of the AAAI Conference on Artificial
  Intelligence, vol~33, pp 7039--7046

\bibitem[{Zhang et~al.(2019)Zhang, Deng, Sun, Chen, Zhang, and
  Chen}]{zhang2019transfer}
Zhang N, Deng S, Sun Z, Chen J, Zhang W, Chen H (2019) Transfer learning for
  relation extraction via relation-gated adversarial learning. arXiv preprint
  arXiv:190808507

\bibitem[{Sahu et~al.(2019)Sahu, Christopoulou, Miwa, and
  Ananiadou}]{sahu2019inter}
Sahu SK, Christopoulou F, Miwa M, Ananiadou S (2019) Inter-sentence relation
  extraction with document-level graph convolutional neural network. arXiv
  preprint arXiv:190604684

\bibitem[{Guo et~al.(2019)Guo, Zhang, and Lu}]{Guo2019}
Guo Z, Zhang Y, Lu W (2019) {Attention Guided Graph Convolutional Networks for
  Relation Extraction} pp 241--251, \doi{10.18653/v1/p19-1024},
  \eprint{1906.07510}

\bibitem[{Zhang et~al.(2019)Zhang, Qi, and Manning}]{Zhang2019}
Zhang Y, Qi P, Manning CD (2019) {Graph Convolution over Pruned Dependency
  Trees Improves Relation Extraction} (2005):2205--2215,
  \doi{10.18653/v1/d18-1244}, \eprint{1809.10185}

\end{thebibliography}

%
%

\end{document}